\documentclass[3p]{elsarticle}

\usepackage{soul}

\usepackage{lineno,hyperref}

\usepackage{times}  %Required
\usepackage{helvet}  %Required
\usepackage{courier}  %Required
\usepackage{url}  %Required
\usepackage{graphicx}  %Required

\usepackage{comment}
\usepackage{subcaption}
\usepackage{todonotes}
\usepackage{multirow}
\usepackage{array}
\usepackage{url}

\usepackage{xspace}
\usepackage{amsfonts}
\usepackage{amssymb,amsmath,amsthm}
\usepackage{mathtools}

\usepackage{colortbl}

\newtheorem{lemma}{Lemma}

\newtheorem{proposition}{Proposition}

\newcommand{\QED}{\hfill\rule[0.2ex]{1.5ex}{1.5ex}}

\newcommand{\bgeq}{\begin{equation}}
\newcommand{\eneq}{\end{equation}}

\newcommand{\reals}[0]{\mathbb{R}}
\newcommand{\mrm}[1]{\mathrm{#1}}

\newcommand{\bm}[1]{\mathbf{#1}}
\newcommand{\bms}[1]{\boldsymbol{#1}}

\newcommand{\tensor}[1]{\bms{\mathcal{#1}}}
\newcommand{\tfold}[2]{\bm{#1}_{(#2)}}
\newcommand{\fslice}[2]{\bm{#1}_{#2}}
\newcommand{\tmmul}[1]{\times_{#1}}
\newcommand{\tvec}[1]{\mathrm{vec}\left(#1\right)}

\newcommand{\norm}[1]{\| #1 \|}
\newcommand{\fnorm}[1]{\norm{#1}_{F}}
\newcommand{\argmin}{\mathrm{argmin}}
\newcommand{\pinv}[1]{#1^{+}}

\newcommand{\mseq}[2]{\bm{#1}^{(#2)}}

\newcommand{\fb}{FB}
\newcommand{\te}[1]{\mathcal{#1}}

\newcommand{\fp}{fact prediction}
\newcommand{\lr}{link ranking}

\newcolumntype{P}[1]{>{\centering\arraybackslash}p{#1}}
\newcolumntype{M}[1]{>{\centering\arraybackslash}m{#1}}
\newcommand*\bigdot{\mathpalette\bigcdot@{.5}}

\journal{Journal of Web Semantics}

\newcommand{\mycomment}[1]{}

\begin{document}

\begin{frontmatter}

\title{Knowledge Graph Fact Prediction via Knowledge-Enriched Tensor Factorization}

\author{Ankur Padia, Konstantinos Kalpakis, Francis Ferraro and Tim Finin}
\address{\{ankurpadia, kalpakis, ferraro, finin\}@umbc.edu \\ University of Maryland, Baltimore County\\ Baltimore, MD, USA}

\begin{abstract}
We present a family of novel methods for embedding knowledge graphs into real-valued tensors. These tensor-based embeddings capture the ordered relations that are typical in the knowledge graphs represented by semantic web languages like RDF. Unlike many previous models, our methods can easily use prior background knowledge provided by users or extracted automatically from existing knowledge graphs. In addition to providing more robust methods for knowledge graph embedding, we provide a provably-convergent, \textit{linear} tensor factorization algorithm. We demonstrate the efficacy of our models for the task of \textit{predicting new facts} across eight different knowledge graphs, achieving between 5\% and 50\% relative improvement over existing state-of-the-art knowledge graph embedding techniques. Our empirical evaluation shows that all of the tensor decomposition models perform well when the average degree of an entity in a graph is high, with constraint-based models doing better on graphs with a small number of highly similar relations and regularization-based models dominating for graphs with relations of varying degrees of similarity.

\end{abstract}

\end{frontmatter}

%\linenumbers

\section{Introduction}

Knowledge graphs are gaining popularity due to their effectiveness in supporting a wide range of applications, ranging from speed-reading medical articles via entity-relationship synopses \cite{ernst2015knowlife}, to training classifiers via distant supervision \cite{mintz2009distant}, to representing background knowledge about the world \cite{auer2007dbpedia,bollacker2008freebase}, to sharing linguistic resources \cite{miller1995wordnet}. Large, broad-coverage knowledge graphs like DBpedia, Freebase, Cyc, and Nell \cite{ringler2017one} have been constructed from a combination of human input, structured and semi-structured datasets, and information extraction from text, and further refined by a mixture of machine learning and data analysis algorithms.  While they are immensely useful in their current state, much work remains to be done to detect the many errors they contain and enhance them by adding relations that are missing.  As a simple example, consider instances of the {\it spouse} relation in the DBpedia knowledge graph. This relation holds between two people and is symmetric, yet the DBpedia version from October 2016 has 3,743 relations where one of the entities is not a type of Person in DBpedia's native ontology and more than half of the inverse relations are missing\footnote{These observations were made based on data from SPARQL queries run on the public endpoint \cite{dbpediasparql} in December 2017.}.

One approach to improving a large knowledge graph like DBpedia is to extend and exploit ontological knowledge, perhaps in the form of logical or probabilistic rules.  However, two factors make this approach problematic:  the presence of noise in the initial graphs, and the large size of the underlying ontologies.  For example, in DBpedia it is infeasible to do simple reasoning with property domain and range constraints because the noisy data produces too many contradictions.  The size of DBpedia's schema, with more than 62K properties and 100K types, makes a rule-based approach difficult, if not impossible.

\begin{table}
  \centering
  \renewcommand{\arraystretch}{1.2}
  \resizebox{\textwidth}{!}{
    \begin{tabular}{|c|c|c|c|}
      \hline
      \rowcolor{gray!15} \textbf{Tasks} & \textbf{Alternate terminology} & \textbf{Definition} & \textbf{Example} \\
      \hline
      Link ranking & Link prediction & \textbf{Input}  : Given a relation, r, and an entity e$_{i}$. (e$_{i}$,r, ?) & \textbf{Input} : Where is Statue of Liberty located? \\
      (ranking) & Link recommendation & \textbf{Output} : Rank list of possible entity e$_{j}$ & \textbf{Output} : (1) Germany (2) United States (3) \textbf{New York (city)}\\
      & & \textbf{OR} & (4) New York (state) (5) Brazil \\
      & & \textbf{Input}  : Given a pair of entities, e$_{i}$, and e$_{j}$. (e$_{i}$,?, e$_{j}$) & \\
      & & \textbf{Output} : Rank list of possible relations, \textit{r} & \\
      \hline
      Fact prediction & Link classification & \textbf{Input}   : A triple (a.k.a fact), e$_{i}$, r, and e$_{j}$. & \textbf{Input} : Is the Statue of Liberty located in Germany? \\
      (classification) & Fact classification & \textbf{Output} : 0 (No) or 1 (Yes) & \textbf{Output} : 0 (No) \\
      \hline
    \end{tabular}
  }
\caption{\label{tab:terminilogy} Distinction among various tasks, their definition, alternate terminology, and an example to understand the phrase 'link prediction' and its usage for a given context. Our approach focuses on the 
\textit{Fact Prediction} task, which is a binary classification task.}
\end{table}

Representation learning \cite{bengio2013representation} provides a way to augment or even replace manually constructed ontology axioms and rules. The general idea is to use instances in a large knowledge graph to discover patterns that are common,
and then use these patterns to suggest changes to the graph.  The changes  are often in the form of adding missing types and relations, but can also involve changes to the schema, removing incoherent instances, merging sets of instances that describe the same real-world entity, or adding or adjusting probabilities for relations.  One popular approach for representation learning systems is based on learning how to \textit{embed} the entities and relations in a graph into a real-valued vector space, allowing the entities and relations to be represented by dense, real-valued vectors. The entity and relation embeddings can be learned either independently or jointly, and  then used to predict additional relations that are missing. Jointly learning the embeddings allows each to enhance the other.

\begin{figure}
\centering
\begin{subfigure}{.48\textwidth}
  \centering
  \includegraphics[scale=0.63]{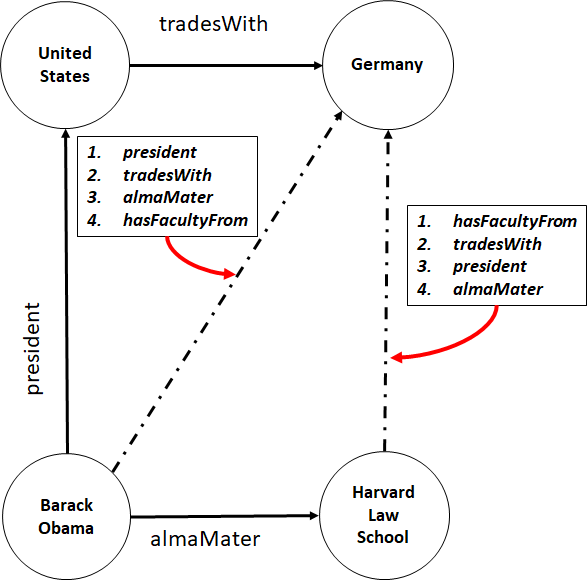}
  \caption{Link Ranking/Link Prediction}
  \label{fig:sub1}
\end{subfigure}%
~
\begin{subfigure}{.48\textwidth}
  \centering
  \includegraphics[scale=0.63]{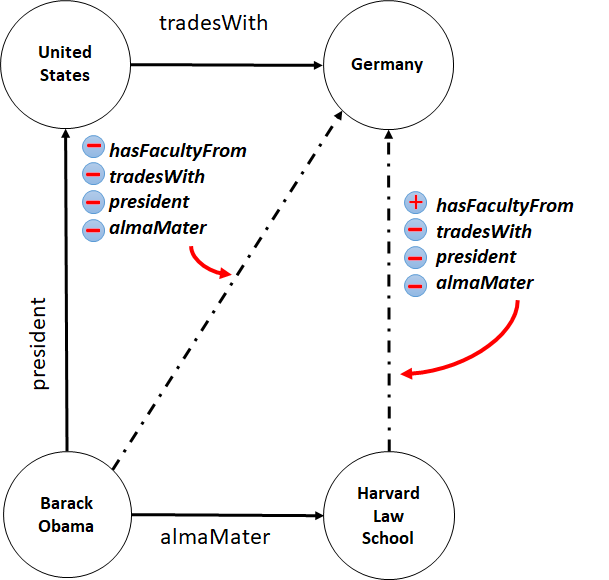}
  \caption{Fact Prediction/Fact classification}
  \label{fig:sub2}
\end{subfigure}
\caption{Link Ranking vs. Fact Prediction. Consider a toy knowledge graph with four entities and four relations. \textbf{Link ranking} aims to rank relations for a given pair of entities and is meaningful in the cases where at least one relation holds between a given pair of entities, e.g., (Barack\_Obama, ?, United\_States) and not (Barack\_Obama, ?, Germany). On the other hand, \textbf{fact prediction} is the task of deciding which relations are likely to hold between a pair of entities. Link ranking (or recommendation) is a ranking problem, while fact prediction is a binary classification problem.}
\label{fig:toyexample}
\end{figure}

Current state-of-the-art systems of this type compute embeddings to support the task at hand, which might be \textit{\lr} (or \textit{link recommendation}), or \textit{\fp} (Table \ref{tab:terminilogy}). Link ranking tries to populate the knowledge graph by recommending a list of relations that could hold between a subject--object pair of entities. It assumes that at least one relation exists between the given pair of entities, and is a ranking problem. On the other hand, fact prediction identifies the correct facts from incorrect ones, and is a binary classification problem. To better understand the difference between link ranking and fact prediction, consider the example shown in Figure \ref{fig:toyexample}. Here solid lines indicate the observed (correct) relations among the entities; dashed lines indicate the relations we are interested in making recommendation for or identifying their correctness. For the pair (Barack Obama, Germany), \textbf{none} of the recommended relations (in the solid box) can hold. However, due to the design of the problem, a \textit{link prediction} system is required to produce a list of potential relations. On the other hand, for the pair (Harvard Law School,Germany), one relation, \textit{hasFacultyFrom}, can hold while the remaining ones cannot.

In the case of fact prediction, we are interested in making a determination (binary classification) whether or not a relation holds between a given pair of entities. 
Fact prediction is an important task, as models for it can help identify erroneous facts present in a knowledge graph and also filter facts generated by an  inference system or information extraction system. As shown in Figure \ref{fig:toyexample} the circumscribed minus sign (``-'') indicates that the relation cannot hold and circumscribed plus sign (``+'') that it may hold. Fact prediction can be used as an pre- and post- processing step to link prediction.

Many previous systems have attacked the \lr{} task (see Section \ref{related_work}), which involves finding, scoring and ranking links that could hold between a pair of entities. Having such a ranked list is useful and could support, for example, a system that showed the list to a person and asked her to check off the ones that hold.  The results of the \lr{} task can also be used to predict facts that do hold between a pair of entities, of course. But it introduces the need to learn good thresholds for the scores to separate the possible from the likely. Achieving high accuracy may require that the thresholds differ from one relation to another. Thus we have a new problem that we need to train a system to solve -- learning optimal thresholds for the relations. Since we are only interested in extending a knowledge graph with relations that are likely to hold (what we call facts), our approach is designed to solve it directly.   Thus we have the \fp{} task: given a knowledge graph, learn a model that can classify relation instances that are very likely to hold. This  task is more specific than \lr{} and more directly solves an important problem.

Embedding entities and relations into a vector space has been shown to achieve state-of-the-art results. Such embeddings can be generated using tensor factorization or neural network based approaches. Tensor--based approaches like RESCAL \cite{Nickel2011} jointly learn the latent representation of entities and relations by factorizing the tensor representation of the knowledge graph. Such a tensor factorization could be further improved by imposing constraints, such as non-negativity on the factors, to achieve better sparsity and prediction performance. Moreover, tensor factorization methods, like Tucker and Canonical Polyadic (CP) decompositions \cite{Kolda2009}, have also been applied to knowledge graphs to obtain ranking of facts~\cite{Franz2009}. RESCAL and its variants~\cite{Krompas2015,krompass2013non} have achieved state-of-the-art results in predicting missing relations on real-world knowledge graphs. However, such extensions require additional  schema information, which may be absent or require significant effort to provide. 

Neural network based approaches, like TransE \cite{bordes2013translating} and DistMult \cite{yang2014embedding}, learn an embedding of a knowledge graph by minimizing the ranking loss. As a result, they learn representations in which likely links are ranked higher than unlikely ones.  These are evaluated with Mean Reciprocal Rank, which emphasizes the ordering or ranking of the candidate links rather than their correctness.  DistMult further assumes that each relation is symmetric.  CompleEx \cite{trouillon2016complex} relaxes the assumption of symmetric relations by representing the embedding in a vector space of complex, rather than real, numbers. DistMult and ComplEx have both been shown to yield state-of-the-art performance.

However, these models \cite{Nickel2011,krompass2013non,bordes2013translating,trouillon2016complex,yang2014embedding} do not explicitly exploit the similarity among the relations when computing entity and relation embeddings, nor have they studied the role that relation similarities have on regularizing and constraining the underlying relation embeddings and the effect on performance in fact prediction task. Other more distantly related methods \cite{demeester2016lifted,minervini2017adversarial,lengerich2017retrofitting} attempt to learn entity and relation embeddings using association among the relations but as described in the Section \ref{related_work}, the approaches need external text sources to determine the association among the relations/predicates and hence are not standalone like ours. The closest approach which does not depend on an external source (i.e., is \textit{standalone}) is ~\citet{minervini2017regularizing}, which uses limited relation similarity cases (i.e., {\tt{inverse}} and {\tt{equivalence}}).  This can be easily modeled by our approach using weighted regularization and hence the regularization provided by \cite{minervini2017regularizing} is a special case of our regularization approach.

Our work addresses these deficiencies and make three contributions.  First, we develop a framework to learn entity and relation embeddings that incorporates similarity among the relations as prior knowledge. This framework allows us to both generalize existing work \citep{padia2016inferring} and provide \textbf{three novel embedding methods}. Our models are based on the intuition that the importance of relations varies in predicting missing relations in a given multi-relational dataset, e.g., knowing that someone is a country's President greatly increases the chances of being a citizen of the country. Formally, each method optimizes an augmented reconstruction loss objective (Section \ref{methodology}) Additionally, we use Alternate Least Squares instead of gradient descent to solve the resulting optimization problems.

Second, we evaluate each model, comparing it to state-of-the-art tensor decomposition models (RESCAL and its non-negative variant Non-negative RESCAL) on \textbf{eight real-world datasets/knowledge graphs} on fact prediction task. These datasets exhibit varying degrees of similarity among the relations, allowing us to study our framework's efficacy in varying settings. We provide insight into our models and shed light on the effect of similarity regularization on on the quality of learned embedding for the task and describe how the embedding changed with varying graph sparsity. We show that the quadratic model perform well in general and in most cases, embedding using our quadratic+constraint model perform the best. We also consider our models as \textit{one-best fact prediction} systems, allowing us to compare against TransE, and popular benchmarks DistMult, and ComplEx. Our methods yield \textbf{consistent relative improvements of more than 20\%} over these baselines, while having the same asymptotic time complexity. 

Finally, we make a theoretical contribution by providing a \textbf{provably convergent} factorization algorithm that matches, and often outperforms, the baselines. We also empirically investigate its convergence on two standard datasets.

\section{Related work}
\label{related_work}

Significant work has been done over the past decades on methods for improving a given knowledge graph by identifying likely errors and either correcting or removing them and by predicting additional facts or relations and adding them to the graph.  Paulheim \cite{paulheim2017knowledge} provides an overview of techniques for these tasks, which he calls knowledge graph refinement. Our interest is in the subset of this general problem that involves using embeddings to identify the correct relations between pairs of entities already in a knowledge graph as opposed to link prediction or recommendation, 

Knowledge graph embeddings can be created using tensor factorization or neural network based approaches. Both aim to learn a scoring function which assigns a score to a triple, $(s,r,o)$ where $s$ is the subject, $r$ is the relation and $o$ is the object. They learn an embedding using a combination of techniques including the use of regularization, constraints or external information. The choice of the techniques affects both the embedding and the types of applications for which they are suited. We describe a few of them and additional details can be found in \cite{nickel2016review}.

\textbf{Neural network based approaches.} Neural network methods like TransE~\cite{bordes-mlj13} and Neural Tensor Network (NTN)~\cite{socher2013reasoning} embed the entities and relations present in multi-relational data using marginal loss. The embeddings are learned in a manner that ranks correct  (i.e., positive) triples higher than incorrect (i.e., negative triples). For each triple $(s,r,o)$, TransE tries to bring the object $o$ closer to the sum of subject $s$ and relation $r$ with a linear scoring function $||\textbf{s}+\textbf{r}-\textbf{o}||$. NTN, on the other hand, uses the combination of a bilinear model ($\textbf{s}^T\textbf{W}_r\textbf{o}$) and a linear one ($\textbf{W}_{rs}\textbf{s}+\textbf{W}_{ro}\textbf{o}+\textbf{b}_r$) where $\textbf{W}_{rs}, \textbf{W}_{ro}$, and $\textbf{W}_{r}$ are the relation embeddings. NTN has more parameters than  TransE, making it generally more expressive.

TransE's approach was extended by TransH~\cite{wang2014knowledge}, which projects relations in a hyperplane with a translation operation on the hyperplane. Subsequently, DistMult \cite{yang2014embedding} and ComplEx~\cite{trouillon2016complex} have been shown to learn better embeddings and perform better then TransE, TransH and NTN, achieving what are currently considered to be state-of-the-art results. DistMult is a simpler version of RESCAL where the relation embedding matrix is assumed to be diagonal. However, since its scoring function is symmetric, it considers each relation to be symmetric, and consequently cannot distinguish the difference between the subject and object. This is a serious drawback in domains with asymmetric relations (e.g., $hasParent$, $attacks$, $worksFor$). ComplEx uses the same number of parameters as DistMult and overcomes this drawback by embedding relations in the vector space of complex numbers, so that each relation's embedding vector has a real and an imaginary part. ComplEx uses the both the real and imaginary parts of subject, predicate, and object embeddings to compute the score.

HoLE~\cite{nickel2016holographic} learns entity and relation embeddings to compute a triple's score with fewer parameters than RESCAL. However, since \cite{hayashi2017equivalence} showed that the holographic embeddings are isomorphic to those of ComplEx, we limit our focus on  DistMult and ComplEx. An approach from \citet{guo2018knowledge} learns embeddings using ComplEx's objective function and iteratively modifies them using rules learned with AMIE \cite{galarraga2013amie}. Such rules can be converted to corresponding score values as entries for the similarity matrix used in our approach (Section \ref{slice_sim_matrix}), using a function like Equation 6 in \citet{guo2018knowledge}. As the number of atoms in a rule can vary, engineering a function to compute a score for a variable length rule and understanding its effect on fact prediction task requires exploration; we leave this for future work. We compare the quality of our embedding with those of the frequently used baseline approaches DistMult and ComplEx and achieve significant improvement on the fact prediction task.  

\textbf{Tensor factorization based approaches.} These approaches compute embeddings by factorizing a knowledge graph's tensor and using the learned factors to assigns a score to each triple. Scores can be boolean, reals, or non-negative reals  depending on the factorization constraints. Boolean Tensor Factorization (BTF) \cite{miettinen2011boolean} decomposes an input tensor into multiple binary-valued factor tensors. The value of the input tensor is reconstructed using boolean operators on the corresponding individual values of the tensor factors. BTF was extended in \cite{erdos2013discovering} by incorporating a Tucker tensor decomposition \cite{Kolda2009} to predicts links. Each factor contains a boolean value, but since the learned values are boolean, the predicted values are constrained to be either 0 or 1. In contrast, our model assigns a real number to each possible link. 

Methods like RESCAL~\cite{Nickel2011} and its schema-based extension \cite{Krompas2015} decompose a tensor into a shared factor matrix and a shared compact factor tensor \cite{Nickel2012}. To better model protein interaction networks and social network data, \Citet{krompass2013non} imposed non-negativity constraints on these factors, but as we show empirically in Section \ref{sec:timecomplexity}, doing so increases the running time of the factorization and introduces scalability issues. Other examples of utilizing schema information include \citet{Krompas2015}, who use schema information to decompose a tensor using type constraints and updates the factor values following a relation's \textit{rdfs:domain} and \textit{rdfs:range}, and \citet{minervini2016latent}, who incorporate schema information in latent factor models to improve the link prediction task.  All of the proposed extensions seem to work well only when the average degree of the entities is high or all of the relations are equally important in predicting the correctness of (possible) facts. Finally, while these approaches offer empirical evidence for the convergence of their iterative algorithms, no convergence guarantees or analysis are available.

Work that can be considered close to ours is \citet{minervini2017regularizing}, which requires pre-defined equivalence and inverse properties on relations. In contrast, we use a data-driven and self-contained approach an do not rely on or require a schema, pre-trained embeddings or external text corpus. Their approach uses two formulations: one in which the equivalences define hard constraints and another in with soft constraints. While the soft constraints take the same form as the relation regularization we use (i.e., Frobenius between relation embeddings). Our approach is supported by the intuition that not all relations participate equally to identify the fact, which provides more flexibility by weighting different relations. Due to this flexibility \cite{minervini2017regularizing} can be considered as a special case of our approach to provide regularization described here and in our preliminary work \cite{padia2016inferring}. Additionally, we do not require inclusion of a rich semantic schema. In the absence of a schema (i.e., without using regularization via equivalence or inverse axioms), their approach reduces to the that of TransE, DistMult, and ComplEx, with which we compare our approach in Section~\ref{sec:transE}. 

\textbf{More distantly related work.} There are approaches that use external information, either from a text corpus or pre-trained embeddings, to regularized knowledge graph embeddings for downstream applications. These are somewhat related to our approach, which is data driven, \textit{self contained} and does not rely a corpus or pre-trained embeddings. Our regularization approach could be added to other regularization methods, enforcing similarity between predicates in the embeddings space. We leave analysis of addition of our regularization to distantly related work for future study.

We mention here a few references for completeness. As mentioned before, NTN \cite{socher2013reasoning} uses pre-trained word embeddings to guide the learning of the knowledge graph embeddings with the intuition that if the words are shared among the entity and relations they share the statistical strength. Beside the use of a text corpus, implication rules are also used to guide the embeddings in some systems ~\cite{demeester2016lifted,minervini2017adversarial,lengerich2017retrofitting}. Such implication rules can come from a lexical corpus, such as WordNet or FrameNet, extracted from the knowledge graph itself \cite{han15} or can be manually crafted. However, this may require considerable amount of human effort, depending on the availability of the lexical resources.

\section{Similarity-driven knowledge graph embedding}

\begin{figure}[t]
\centering
\includegraphics[scale=2.0]{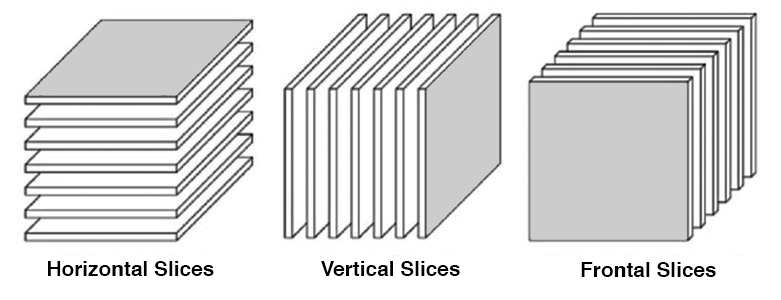}
\caption{The similarity matrix $C$ is used to compute the similarity of pairs of relations in the knowledge graph. Its i$^{th}$ {\em frontal slice} is the adjacency matrix of the i$^{th}$ relation, i.e., a two-dimensional matrix with a row and column for each entity whose values are 1 if the relation holds for a pair and 0 otherwise.}
\label{fig:slices}        
\end{figure}

In this section we first motivate a general \textit{framework} for incorporating existing relational similarity knowledge. We then describe how our three models pre-compute a \textit{similarity matrix} that measures co-occurrence of pairs of relations and use it to regularize or constrain relation embeddings. Two of the three models optimize linear factorization objectives while the third is a robust extension of the quadratic objective described in \citet{padia2016inferring}. 

\subsection{General framework}
Our general framework for similarity-driven knowledge graph embedding relies on minimizing an augmented reconstruction loss. %
The reconstruction objective learns entity and relation embeddings that, when ``combined'' (multiplied), closely approximate the original facts and relation occurrences observed in the knowledge graph. %
We augment the learning process with a \textit{relational similarity} matrix, which provides a holistic judgment of how similar pairs of relations are. %
These similarity scores allow certain constraints to be placed on the learned embeddings; in this way, we allow existing knowledge to enrich the entity and relation embeddings. %

In our framework, we represent a multi-relational knowledge graph of $N_r$ binary relations among $N_e$ entities by the order-3 tensor $\te{X}$ of dimension $N_e \times N_e \times N_r$. %
This binary tensor is often very large and sparse. %
Our goal is to construct dense, informative $p$-dimensional embeddings, where $p$ is much smaller than either the number of entities or the number of relations. %
We represent the collection of $p$-dimensional entity embeddings by $\mathcal{A}$, the collection of relation embeddings by $\mathcal{R}$, and the similarity matrix by $C$. %
The entity embeddings collection $\mathcal{A}$ contains matrices $A_\alpha$ of size $N_e \times p$ while the relation embeddings collection $\mathcal{R}$ contains matrices $\mathbf{R}_k$ of size $p \times p$. %
Recall that the frontal slice $\boldsymbol{X}_{k}$ of tensor $\te{X}$ is the adjacency matrix of the k$^{th}$ binary relation, as shown in Figure \ref{fig:slices}. %
We use $\mathbf{A} \otimes \mathbf{B}$ to denote the Kronecker product of two matrices \textbf{A} and \textbf{B}, vec $\left( \textbf{B}\right)$ to denote the vectorization of a matrix \textbf{B}, and a lower italic letter like \textit{a} to denote a scalar.\footnote{ %
  We use the standard tensor notations and definitions in \citet{Kolda2009}. %
  Recall that the Kronecker product $\mathbf{A} \otimes \mathbf{B}$ of an $(m_1,n_1)$ matrix $\mathbf{A}$ and a $(m_2,n_2)$ matrix $\mathbf{B}$ returns an $(m_1m_2, n_1n_2)$ block matrix, where each element of $\mathbf{A}$ scales the entire matrix $\mathbf{B}$. %
} %

Mathematically, our objective is to reconstruct each of the $k$ relation slices of $\mathcal{X}$, $\mathbf{X}_k$, as the product
\begin{equation}
\mathbf{X}_k \approx \mathcal{A}_\alpha \mathbf{R}_k \mathcal{A}_\beta^\intercal. 
\label{eqn:gen-prod}
\end{equation}
Recall that both $\mathcal{A}_\alpha$ and $\mathcal{A}_\beta$ are matrices: each row is the embedding of an entity. %
By changing the exact form of $\mathcal{A}$---that is, the number of different entity matrices, or the different ways to index $\mathcal{A}$---we can then arrive at different models. %
These model variants encapsulate both mathematical and philosophical differences. %
In this paper, we specifically study two cases. %
First, we examine the case of having only a single entity embedding matrix, represented as $\mathbf{A}$---that is, $\mathcal{A}_\alpha = \mathcal{A}_\beta = \mathbf{A}$. %
This results in a quadratic reconstruction problem, as we approximate $\mathbf{X}_k \approx \mathbf{A} \mathbf{R}_k \mathbf{A}^\intercal.$ %
Second, we examine the case of having two separate entity embedding matrices, represented as $\mathbf{A}_1$ and $\mathbf{A}_2$. %
This results in a reconstruction problem that is linear in the entity embeddings, as we approximate $\mathbf{X}_k \approx \mathbf{A}_1 \mathbf{R}_k \mathbf{A}_2^\intercal.$ %

We learn $\mathcal{A}_\alpha, \mathcal{A}_\beta,$ and $\mathcal{R}$ by minimizing the augmented reconstruction loss
\begin{equation}
\displaystyle
\min_{\mathcal{A}, \mathcal{R}}
\underbrace{f(\mathcal{A}, \mathcal{R})}_{\text{\makebox[0pt]{reconstruction loss}}} +
\overbrace{g(\mathcal{A}, \mathcal{R})}^{\text{\makebox[0pt]{numerical regularization of the embeddings}}} +
\underbrace{f_s(\mathcal{A}, \mathcal{R}, C)}_{\text{\makebox[0pt]{knowledge-directed enrichment}}}.
\label{eqn:gen-objective}
\end{equation}
The first term of \eqref{eqn:gen-objective} reflects each of the $k$ relational criteria given by \eqref{eqn:gen-prod}. %
The second term employs standard numerical regularization of the embeddings, such as Frobenius minimization, that enhances the algorithm's numerical stability and supports the interpretability of the resulting embeddings. %
The third term uses our similarity matrix $C$ to enrich the learning process with our extra knowledge.

We first discuss how we construct the similarity matrix $C$ in Section \ref{slice_sim_matrix} and then, starting in Section \ref{methodology}, describe how the framework readily yields three novel embedding models, while also generalizing prior efforts.
Throughout, we show how $C$ can be used as a second type of \textit{regularizer} on the relation embeddings (penalizing large differences in similar relations), or as a \textit{constraint} that forces embeddings of similar relations to be near one another and dissimilar relations to be further apart. %
In particular, we demonstrate that when using
\begin{enumerate}
\item a linear objective with $C$ as a \textit{regularizer} (Sect.\ \ref{sec:model:linear_regularized}), we obtain a competitive, provably convergent algorithm; %
\item a quadratic objective with $C$ as a \textit{constraint} (Sect.\ \ref{sec:model:quad_constraint}), we obtain a method that relies on the well-known quadratic form, while resulting in significantly higher performance.
\end{enumerate}

\subsection{Slice similarity matrix: $\boldsymbol{C}$}
\label{slice_sim_matrix}

Each element of the $N_r \times N_r$ matrix $\boldsymbol{C}$ represents the similarity between a pair of relations, i.e., frontal tensor slices $\boldsymbol{X}_{i}$ and $\boldsymbol{X}_{j}$, and is computed using the following equation:
  \begin{eqnarray}
  \label{eq:heuristic_function}
  (Symmetric)\quad C_{i,j} & = & \frac{|(S(X_{i})\cup O(X_{i}))\cap(S(X_{j})\cup O(X_{j}))|}{|(S(X_{i})\cup O(X_{i}))\cup(S(X_{j})\cup O(X_{j}))|}\forall1\leq i,j\leq N_{r}
  \end{eqnarray}

where $S(\boldsymbol{X}_i)$ is the set of subjects of the matrix $\boldsymbol{X}$ holding the $i^{th}$ relation, and similarly for the object $O(\boldsymbol{X}_i)$. $|S(\boldsymbol{X})|$ gives the cardinality of the set.%
Intuitively, we measure similarity of two relations using the overlap in the entities observed with each relation. Two relations that operate on more of the same entities are more likely to have \textit{some} notion of being similar. %
The numerator equals the number of common entity pairs present across the two frontal slices (relations), while the denominator is used to normalize the score between zero and one. Beside Equation \ref{eq:heuristic_function} we also consider several other similarity function:

\begin{eqnarray}
\label{eq:agency}
\textit{(Agency)}\quad C & = & \frac{|S(X_{i})\cap S(X_{j})|}{|S(X_{i})\cup S(X_{j})|}\forall1\leq i,j\leq N_{r}
\end{eqnarray}
\begin{eqnarray}
\label{eq:patient}
\textit{(Patient)}\quad C_{i,j} & = & \frac{|O(X_{i})\cap O(X_{j})|}{|O(X_{i})\cup O(X_{j})|}\forall1\leq i,j\leq N_{r}
\end{eqnarray}
\begin{eqnarray}
\label{eq:transitivity}
\textit{(Transitivity)}\quad C_{i,j} & = & \frac{|S(X_{i})\cap O(X_{j})|}{|S(X_{i})\cup O(X_{j})|}\forall1\leq i,j\leq N_{r}
\end{eqnarray}
\begin{eqnarray}
\label{eq:reverse_transitivity}
\textit{(Reverse Transitivity)}\quad C_{i,j} & = & \frac{|O(X_{i})\cap S(X_{j})|}{|O(X_{i})\cup S(X_{j})|}\forall1\leq i,j\leq N_{r}
\end{eqnarray}

We can view a knowledge graph's nodes and edges as representing a flow of information, with subjects and objects acting as information producers and consumers, respectively. Tensor factorization captures this interaction \cite{Nickel2011}.

We experimented with all of the similarity functions and report the evaluation result in Section \ref{sec:evaluation}. For most of our experiments we used the similarity obtained from transitivity, as we found it gave the best overall performance.

\begin{figure*}[p!] %forces figure onto page by itself. TF: I think these may look batter if larger
\newcommand{\myscale}{0.4}
\centering
\begin{subfigure}{\textwidth}
\centering
\includegraphics[scale=0.48]{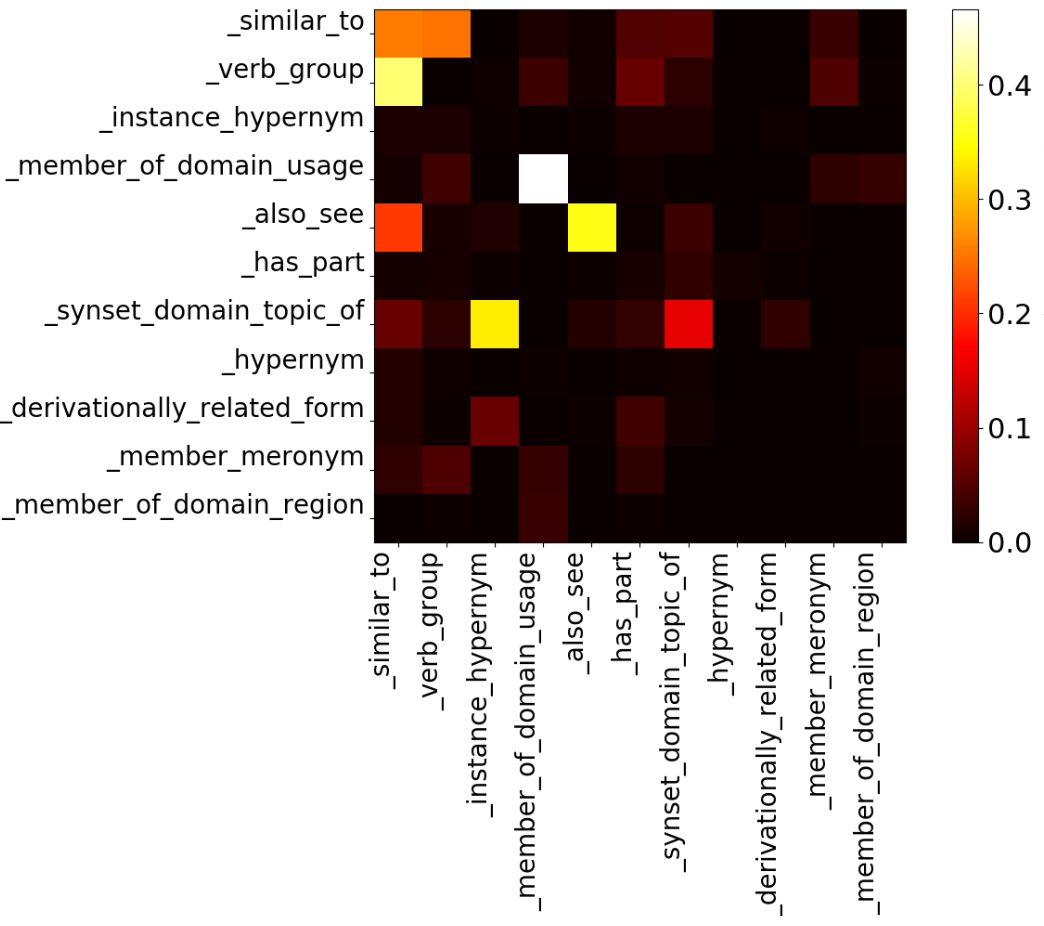}
\caption{The computed similarity matrix for the WordNet (WN18RR) dataset using the transitivity criterion.}
\label{fig:heatmap:wn18}
\end{subfigure}

%~
\begin{subfigure}{\textwidth}
\centering
\hspace{-1cm}
\includegraphics[scale=0.52]{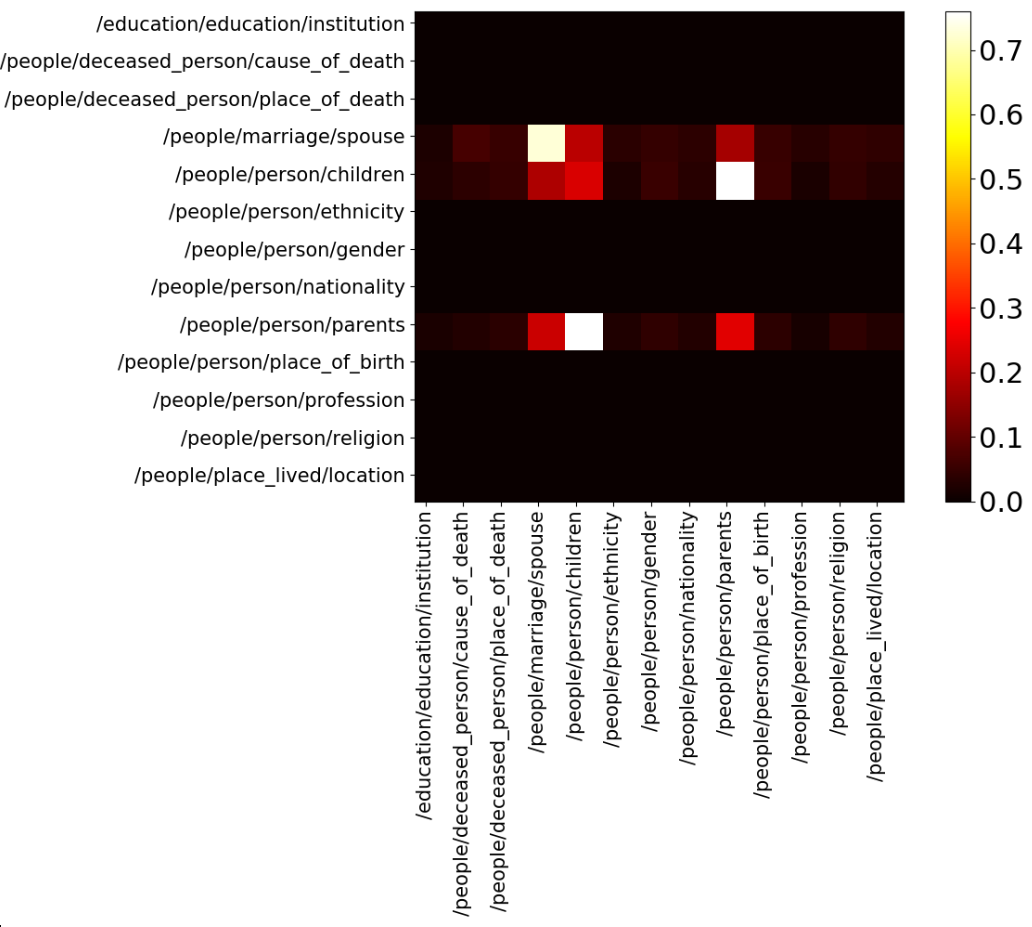}
\caption{The computed similarity matrix for the Freebase (FB13) dataset using the transitivity criterion.}
\label{fig:heatmap:fb13}
\end{subfigure}
\caption{These heatmaps visualize the similarity among the relations present in the knowledge graph for the WN18RR and FB13 datasets. More darkly colored cells represent lower similarity and brighter ones indicate higher similarity.}
\label{fig:heatmap} 
\end{figure*}

Our similarity function in Eq. \ref{eq:heuristic_function} is symmetric. An asymmetric similarity function, like the Tversky index \cite{gawron2014improving}, could be used, but we found its performance to be comparable to our simpler symmetric similarity function on the \lr task. Figure \ref{fig:heatmap} shows the computed similarity matrices for two of our datasets, WordNet and Freebase, with detailed discussion given in Section \ref{sec:datasets}). 

\subsection{Model 1: \textbf{Linear + Regularized}}
\label{methodology}
\label{sec:model:linear_regularized}

This section presents a linear objective function that can be viewed as a longitudinal extension of previous work that focused on quadratic objectives \citep{padia2016inferring}. We solve the following regularized minimization problem:
\begin{eqnarray}
\label{eq:modifiedrescal+}
\displaystyle
\min_{\mathbf{A_{1}},\mathbf{A_{2}},\textbf{R}_k}
	& f\left(\mathbf{A_{1}},\mathbf{A_{2}},\textbf{R}_k\right)
    + g\left(\mathbf{A_{1}},\mathbf{A_{2}},\textbf{R}_k\right) 
 	+ f_{s}\left(\textbf{C},\textbf{R}_k\right) 
    + f_{\rho}\left(\mathbf{A_{1}},\mathbf{A_{2}},\mathbf{R}_k\right) %
\end{eqnarray}
where we have decomposed the knowledge-directed enrichment term of \eqref{eqn:gen-objective} into two separate terms, $f_s$ and $f_\rho$. Specifically, we minimize
\begin{gather}
f\left(\textbf{\ensuremath{\mathbf{A_{1}}},\ensuremath{\mathbf{A_{2}}},}\textbf{R}_{k}\right) = \frac{1}{2}\left(\sum_{k}||\textbf{X}_{k}-\mathbf{A_{1}}\textbf{R}_k\mathbf{A_{2}^{T}}||_{F}^{2}\right) \\
g\left(\textbf{\ensuremath{\mathbf{A_{1}}},A,}\textbf{R}_{k}\right) = \frac{1}{2}\left(\lambda_{A}||\mathbf{A_{1}}||_{F}^{2}+\lambda_{A}||\mathbf{A_{2}}||_{F}^{2}\right)+\frac{1}{2}\lambda_{e}|| \mathbf{A}_1-\mathbf{A}_2||_{F}^{2} +\frac{1}{2}\left(\lambda_{r}\sum_{k} ||\textbf{R}_{k}||_{F}^{2}\right) \\ f_{s}\left(\textbf{C},\textbf{R}_{k}\right) = \frac{1}{2}\lambda_{s}\sum_{i}\textbf{C}_{k,i}\cdot||\textbf{R}_{k}-\textbf{R}_{i}||_{F}^{2}
\ \ \ \ \ \ \ \forall1\leq i\leq N_{r},1\leq k\leq N_{r} \\ %
f_{\rho}\left(\mathbf{A}_1,\mathbf{A}_2,\mathbf{R}_k\right) = \frac{1}{\rho}\left(||\mathbf{A}_1||_{F}^{2}+||\mathbf{A}_{2}||_{F}^{2} + ||\mathbf{R}_{k}||_{F}^{2} \right) \label{eq:rho_equation}
\end{gather}

\noindent  
Each row $i$ of the matrices \textbf{A$_1$} and \textbf{A$_2$} is a latent representation of the corresponding $i$th entity. The frontal slice $\textbf{R}_k$ is a $p \times p$ matrix representing the interaction of all entities with respect to the k$^{th}$ relationship. The precomputed matrix \textbf{C} is an $N_r \times N_r$ similarity matrix where each element is a similarity score between two tensor slices (relations). The model's objective is to factorize a given data tensor $\te{X}$ into shared matrices \textbf{A$_1$} and \textbf{A$_2$}, and a tensor of relatively low dimension, $\te{R}$, while considering the similarity values present in the matrix \textbf{C}.

In the objective function above, the first term $\mathit{f\left(\textbf{A$_1$}, \textbf{A$_2$},\textbf{R}_k\right)}$ forces the reconstruction to be similar to the original tensor $\te{X}$. The second term, $\mathit{g\left(\textbf{A$_1$},\textbf{A$_2$},\textbf{R}_k\right)}$, is a regularization term to avoid overfitting and nudge \textbf{A$_1$} and \textbf{A$_2$} to be equal. The Frobenius norm $||\cdot||_{F}$ promotes solutions with a small total magnitude, in the sense of Euclidean length.

The third term, $\mathit{f_{s}\left(\textbf{C},\textbf{R}_k\right)}$, provides the longitudinal extension to tensor decomposition. It supports the differential contribution of tensor slices in the reconstruction of the tensor $\te{X}$. The similarity values ($\mathit{C}_{i,j}$) force slices of the relational tensor to decrease their differences between one another. 
To reduce the degree of entity embeddings from quadratic to linear, we use the split-variable technique by replacing variable \textbf{A} with two variables, \textbf{A$_1$} and \textbf{A$_2$}, that are constrained to be equal. To guarantee  convergence, we add an additional term, $f_\rho$ (Equation \ref{eq:rho_equation}), to Equation \ref{eq:modifiedrescal+} which has partial Hessians that are positive definite. The use of $\rho$ is motivated at high-level from proximal algorithms \cite{parikh2014proximal} to ensure the strict convexity of the objective function, as described in \ref{sec:appendix}).  

\subsubsection{Computing factor matrices $\mathbf{A_{1}}$, $\mathbf{A_{2}}$ and $\mathbf{R}$$_k$}

We compute the factor matrices with alternating least squares (ALS)~\cite{Bader2007}, a non-linear block Gauss-Seidel method in which the blocks are the unknowns $\mathbf{A_{1}}$, $\mathbf{A_{2}}$ and the frontal slices of $\tensor{R}$. We consider the partial objective functions that need to be optimized, one for each block when the other objective function blocks are kept fixed. We find that each function is a quadratic form of the unknown block whose Hessian is always positive semi-definite, and it is positive definite whenever $\rho > 0$. In other words, each partial objective function is a strictly convex quadratic (least squares) problem with a unique global minimum. In particular, taking the gradient of Eq. \ref{eq:modifiedrescal+} with respect to $\mathbf{A_{1}}$ and setting it equal to zero, we obtain the update rule for $\mathbf{A_{1}}$. 

	\hspace{-4em}
	\begin{eqnarray}
	\mathbf{A_{1}} & \leftarrow &  \left[\sum_{k=1}^{N_r}\mathbf{X}_{k}\mathbf{A_{2}}\mathbf{R}_{k}^{T} + \lambda_{A}\mathbf{A_{2}}\right]\left[\sum_{k=1}^{N_r}\mathbf{R}_{k}\mathbf{A_{2}}^{T}\mathbf{A_{2}}\mathbf{R}_{k}^{T}+\left(\lambda_{A} + \lambda_{e} + \frac{1}{\rho}\right)\mathbf{I}\right]^{-1}
	\end{eqnarray}
    
\noindent
Similarly, taking the gradient of Eq. \ref{eq:modifiedrescal+} with respect to $\mathbf{A_{2}}$ and setting it equal to zero, we obtain the update rule for $\mathbf{A_{2}}$.
	\hspace{-4em}
	\begin{eqnarray}
	\mathbf{A_{2}} & \leftarrow & \left[\sum_{k=1}^{N_r}\mathbf{X}_{k}^{T}\mathbf{A_{1}}\mathbf{R}_{k} + \lambda_{A}\mathbf{A_{1}}\right]\left[\sum_{k=1}^{N_r}\mathbf{R}_{k}^{T}\mathbf{A_{1}}^{T}\mathbf{A_{1}}\mathbf{R}_{k}+\left(\lambda_{A} + \lambda_{e} + \frac{1}{\rho}\right)\mathbf{I}\right]^{-1}
	\end{eqnarray}
\noindent

The unknown matrix \textbf{R}$_k$ can be found 
by solving following variant of Eq. \ref{eq:modifiedrescal+}, which is a ridge regression problem with positive definite Hessian.
    \hspace{-4em}
	\begin{eqnarray}
	\nonumber
        \min_{vec(\mathbf{R}_k)}
        ||vec(\mathbf{X}_{k})-(\mathbf{A_{2}}\otimes\mathbf{A_{1}})vec(\mathbf{R}_{k})||^2+\left(\lambda_{r} + \frac{1}{\rho}\right)||vec(\mathbf{R}_{k})||^2 + \lambda_{s}\sum_{i}  ||vec(\mathbf{R}_{k}-\mathbf{R}_{i})||^2
    \end{eqnarray}

Since the problem is strictly convex, the unique minimum is obtained by setting the gradient to $0$, leading to the following update rule for \textbf{R}$_k$.
\begin{eqnarray}
%\hspace*{-15cm}
\mathbf{R}_{k} &\leftarrow& \left(\left(\mathbf{A_{2}}\otimes\mathbf{A_{1}}\right)^{T}\left(\mathbf{A_{2}}\otimes\mathbf{A_{1}}\right)+\left(\lambda_{r} + \frac{1}{\rho}\right)\mathbf{I}+(\lambda_{s}\sum_{i}^{N_r}\mathbf{C}(k,i))\mathbf{I}\right)^{-1}\left(\mathbf{A_{2}}\otimes\mathbf{A_{1}})vec(\mathbf{X}_{k})\right) 
\end{eqnarray}
 
\subsection{Model 2: \textbf{Quadratic + Constraint}}
\label{sec:model:quad_constraint}

In the second model, we consider the decomposition of $\te{X}$ into a compact relational tensor $\te{R}$ and quadratic entity matrix \textbf{A}. We solve the following problem

\begin{eqnarray}
& \displaystyle
  \min_{\mathbf{A}, \mathbf{R}_k}
  f(\mathbf{A},\mathbf{R}_{k})+g(\mathbf{A},\mathbf{R}_{k})
\label{quadConst}
\end{eqnarray}
under the constraint that relations with high similarity are near one another.
\begin{eqnarray}
& ||\mathbf{R}_{i}-\mathbf{R}_{j}||_{F}^{2}=1-C_{ij},1\leq i,j\leq n.
\label{eq:lagConstraint}
\end{eqnarray}
The two terms of our objective are expressed as follows.
\begin{eqnarray}
 & f(\mathbf{A},\mathbf{R}_{k})=\frac{1}{2}{\textstyle {\displaystyle \sum_{k}}}||\mathbf{X}_{k}-\mathbf{AR}_{k}\mathbf{A}^{T}||_{F}^{2}\\
 & g(\mathbf{A},\mathbf{R}_{k})=\frac{1}{2}\lambda_{a}||\mathbf{A}||_{F}^{2}+\frac{1}{2}\lambda_{r}\sum_{k}||\mathbf{R}_{k}||_{F}^{2}
\end{eqnarray}
\noindent
Here \textbf{A} is a \textit{n} $\times$ \textit{p} matrix where each row represents the entity embeddings and \textbf{R}$_k$ is a \textit{p} $\times$ \textit{p} matrix representing the embedding for the \textit{k}$^{th}$ relation capturing the interaction between the entities. The first term \textit{f} forces the reconstruction to be similar to the original tensor and the second regularizes the unknown \textbf{A} and \textbf{R}$_{k}$ to avoid overfitting. In order to incorporate similarity constraints, we modify Eq. \ref{quadConst} to solve the dual objective, via Lagrange multipliers $\lambda_{ij}$ as below.
\begin{gather}
   \displaystyle
  \min_{\mathbf{A}, \mathbf{R}_k}
  f(\mathbf{A},\mathbf{R}_{k})+g(\mathbf{A},\mathbf{R}_{k})+f_{\text{Lag}}(\mathcal{R},\mathbf{C})
\label{quadlag} \\
 f_{\text{Lag}}={\textstyle {\displaystyle \sum_{i}}}{\textstyle {\displaystyle \sum_{j}}}\lambda_{ij}(1-||\mathbf{R}_{i}-\mathbf{R}_{j}||_{F}^{2}+\mathbf{C}_{ij}).
 \label{eq:langQuadConstraint}
\end{gather}
The $f_{\text{lag}}$ term represents the model's knowledge-directed enrichment component.

\subsubsection{Computing Factor Matrices, \textbf{A}, \textbf{R}$_k$ and Lagrange Multipliers $\lambda_{ij}$}
\label{sec:quadLagFactor}

We compute the unknown factor matrices using Adam optimization \cite{kingma2014adam}, an extension to stochastic gradient descent.  Each unknown is updated in the alternative fashion, in which each parameter is updated while treating the others as constants. Each unknown parameter of the model, \textbf{A} and \textbf{R}$_k$, is updated with different learning rate. We empirically found that the error value of the objective function decreases after few iterations. Taking the partial derivative of the Eq. \ref{quadlag} with respect to \textbf{A} and equating to zero we obtain the following update rule for \textbf{A}.
\begin{eqnarray}
&
\mathbf{A}\leftarrow\left(\mathbf{X}_{k}^{T}\mathbf{AR}_{K}+\mathbf{X}_{k}\mathbf{AR}_{k}^{T}\right)\left(\mathbf{R}_{k}^{T}\mathbf{A}^{T}\mathbf{AR}_{k}^{T}+\lambda_{a}\mathbf{I}\right)^{-1}
\end{eqnarray}
\noindent
Since we are indirectly constraining the embeddings of \textbf{A} through slices of the compact relation tensor $\te{R}$, we obtain the same update rule for $\mathbf{A}$ as in RESCAL \citep{Nickel2011}. %\textbf{R}$_k$. 
%Adapting the ASALAN method \cite{}, and
By equating the partial derivatives of Eq. \ref{quadlag}  with respect to the unknowns \textbf{R}$_k$ and  $\lambda_{ij}$ to 0, and solving for those unknowns, we obtain the following updates: %update rules for \textbf{R}$_k$ and Lagrange multipliers $\lambda_{ij}$:
\begin{gather}
vec(\mathbf{R}_{K})\leftarrow\left(\left(\mathbf{A}^{T}\mathbf{A}\otimes \mathbf{A}^{T}\mathbf{A}\right)+\lambda_{r}\mathbf{I}+\lambda_{i=k,j}\mathbf{I}\right)^{-1}\left(\left(\mathbf{A}\otimes \mathbf{A}\right)^{T}vec(\mathbf{X}_{k})+\sum_{j}\lambda_{i=k,j}vec(\mathbf{R}_{j})\right) \\
\lambda_{ij}\leftarrow||\mathbf{R}_{i}-\mathbf{R}_{j}||_{F}^{2}+\mathbf{C}_{ij}-1.
\end{gather}

\subsection{Model 3: \textbf{Linear + Constraint}}

This version combines the previous two models: we examine the linear reconstruction loss of Sect. \ref{sec:model:linear_regularized} with the constraints of Sect. \ref{sec:model:quad_constraint}. 

As before, we split the entity embedding of \textbf{A} into \textbf{A}$_1$ and \textbf{A}$_2$. Additionally, we apply the same constraint as in Eq. \ref{eq:lagConstraint} and solve following constrained problem:
\begin{eqnarray}
\label{eq:linearConstraint}
\displaystyle
\min_{\mathbf{A_{1}},\mathbf{A_{2}},\textbf{R}_k}
	& f\left(\mathbf{A_{1}},\mathbf{A_{2}},\textbf{R}_k\right)+
     g\left(\mathbf{A_{1}},\mathbf{A_{2}},\textbf{R}_k\right) 
\end{eqnarray}
such that,
\begin{eqnarray}
& ||\mathbf{R}_{i}-\mathbf{R}_{j}||_{F}^{2}=1-C_{ij},1\leq i,j\leq n
\label{eq:lagConstraint2}
\end{eqnarray}
where,
\begin{gather}
f\left(\textbf{\ensuremath{\mathbf{A_{1}}},\ensuremath{\mathbf{A_{2}}},}\textbf{R}_{k}\right) = \frac{1}{2}\left(\sum_{k}||\textbf{X}_{k}-\mathbf{A_{1}}\textbf{R}_k\mathbf{A_{2}^{T}}||_{F}^{2}\right)
\label{eq:flinearConstraint} \\
g\left(\textbf{\ensuremath{\mathbf{A_{1}}},A,}\textbf{R}_{k}\right) = \frac{1}{2}\left(\lambda_{A}||\mathbf{A_{1}}||_{F}^{2}+\lambda_{A}||\mathbf{A_{2}}||_{F}^{2}\right)+\frac{1}{2}\lambda_{e}|| \mathbf{A}_1-\mathbf{A}_2||_{F}^{2} +\frac{1}{2}\left(\lambda_{r}\sum_{k} ||\textbf{R}_{k}||_{F}^{2}\right) %\nonumber
\label{eq:glinearConstraint}
\end{gather}
We rewrite the above constrained problem into a unconstrained one using $\lambda_{ij}$ as a Lagrange multiplier as follows.
\begin{eqnarray}
\label{eq:linearLagMult}
\displaystyle
\min_{\mathbf{A_{1}},\mathbf{A_{2}},\textbf{R}_k}
	& f\left(\mathbf{A_{1}},\mathbf{A_{2}},\textbf{R}_k\right)+
     g\left(\mathbf{A_{1}},\mathbf{A_{2}},\textbf{R}_k\right) 
 	 + f_{\text{Lag}}\left(\textbf{R}_k,\textbf{C}\right) 
\end{eqnarray}
where %
$f\left(\mathbf{A_{1}},\mathbf{A_{2}},\textbf{R}_k\right)$, and    $g\left(\mathbf{A_{1}},\mathbf{A_{2}},\textbf{R}_k\right)$ are same as Eq. \ref{eq:flinearConstraint}. $f_{\text{Lag}}$ is the same as Eq. \ref{eq:langQuadConstraint}.

\subsubsection{Computing the unknowns, \textbf{A}, \textbf{R}$_{k}$ and $\lambda_{ij}$}

As in the previous model, we use an Adam optimizer. Taking the derivative of Eq. \ref{eq:linearConstraint} with respect to \textbf{A}$_1$ and \textbf{A}$_2$, respectively and equating to zero, we obtain the following update rule.
\begin{eqnarray}
 & \mathbf{A}_{1}\leftarrow\left(\lambda_{e}\mathbf{A}_{2}+\mathbf{X}_{k}\mathbf{A}_{2}\mathbf{R}_{k}^{T}\right)\left(\mathbf{R}_{k}\mathbf{A}_{2}^{T}\mathbf{A}_{2}\mathbf{R}_{k}+\lambda_{a_{1}}I+\lambda_{e}I\right)^{-1}
\end{eqnarray}
\begin{eqnarray}
 & \mathbf{A}_{2}\leftarrow\left(\lambda_{e}\mathbf{A}_{1}+\mathbf{X}_{k}^{T}\mathbf{A}_{1}\mathbf{R}_{k}\right)\left(\mathbf{R}_{k}^{T}\mathbf{A}_{1}^{T}\mathbf{A}_{1}\mathbf{R}_{k}+\lambda_{a_{2}}I+\lambda_{e}I\right)^{-1}
\end{eqnarray}
Similarly, taking derivate with respect to $k^{th}$ slice of relation, \textbf{R}$_{k}$ yields the following update rule.
\begin{eqnarray}
 & vec(\mathbf{R}_{k})\leftarrow\left(\left(\mathbf{A}_{2}\otimes \mathbf{A}_{1}\right)^{T}vec(\mathbf{X}_{K})-\sum_{i=k,j}\lambda_{kj}vec(\mathbf{R}_{j})\right)\left(\left(\mathbf{A}_{2}^{T}\mathbf{A}_{2}\otimes \mathbf{A}_{1}^{T}\mathbf{A}_{1}\right)+\lambda_{r}\mathbf{I}-\sum_{j}\lambda_{kj}\right)^{-1}
\end{eqnarray}

\section{Experimental evaluation}
\label{sec:evaluation}

We evaluated the performance of the learned entity and relation embeddings on the \textit{\fp} task, which identifies correct triples from incorrect ones, and compared the results against state-of-the-art tensor decomposition techniques and translation methods, like TransE.
We also demonstrate the convergence of our linear model on two standard benchmark datasets. 

We carried out evaluations on eight real-world datasets.  Five have been extensively used previously as benchmark relational datasets: Kinship, UMLS, WordNet (WN18), WordNet Reverse Removed (WN18RR) and Freebase (FB13). We created a sixth dataset, DBpedia-Person (DB10k), to explore how well our approach works on datasets with a larger number of relations. We created a seventh dataset from FrameNet, an ontological and lexical resource \citep{baker1998Framenet}. Finally, we used the \fb15-237 dataset which was based on Freebase to explore how systems work with a relatively larger number of relations.

We compare our models with state-of-the-art tensor decomposition models, RESCAL and its non-negative variant NN-RESCAL, along with two popular benchmarks, DistMult, which consider the relation embedding matrix to be diagonal, and ComplEx, which represent entities and relation in complex vector space.\footnote{We also experimented with other tensor decomposition models \cite{tensorfactorizationlib} like PARAFAC \cite{harshman1994parafac,bro1997parafac} and TUCKER, but the unfolding of the tensors for the larger datasets (WN18 and FB13) required more than 32GB RAM of memory, which we were unable to support on our testbed.}

\subsection{Datasets}
\label{sec:datasets}
\begin{table}[t]
\renewcommand{\arraystretch}{1.4}
\centering 
\caption{%
  Statistics of the eight datasets used in the evaluation experiments.
  The number of facts represents the number of triples.
}
\label{tab:ds_stats}
\vspace{9pt}
\begin{tabular}{|l|c|c|c|c|c|}
\hline
\rowcolor{gray!15} \textbf{Name} & \# Entities ($\boldsymbol{N_e}$) & \# Relations ($\boldsymbol{N_r}$) & \# \textbf{Facts} & \textbf{Avg. Deg.} & \textbf{Graph Density}  \\ \hline
\textbf{Kinship} & 104  & 26     & 10,686 & 102.75 & 0.98798 \\ \hline     
\textbf{UMLS} & 135  & 49     & 6,752 & 50.01 & 0.37048\\ \hline            
\textbf{\fb15-237} & 14,541 & 237    & 310,116 & 21.32 & 0.00147\\ \hline	
\textbf{DB10k} & 4,397 & 140    & 10,000 & 2.27 & 0.00052\\ \hline			
\textbf{FrameNet} & 22,298  & 16 & 62,344 & 2.79 & 0.00013\\ \hline
\textbf{WN18} & 40,943 & 18 & 151,442 & 3.70 & 0.00009\\ \hline            
\textbf{FB13} & 81,061 & 13 & 360,517 & 4.45 & 0.00005 \\ \hline
\textbf{WN18RR} & 40,943 & 11    & 93,003 & 2.27 & 0.00005\\ \hline	
\end{tabular}
\end{table}

Table \ref{tab:ds_stats} summarizes the key statistics of the datasets: the number of entities ($N_e$), relations ($N_r$) and facts (non-zero entries in the tensor), the average degree of entities across all relations (the ratio of facts to entities) and the graph density (the number of facts divided by square of the number of entities). Note that a smaller average degree or graph density indicates that the knowledge graph is sparser. 

\textbf{Kinship} \cite{Kemp2006} is dataset with information about complex relational structure among 104 members of a tribe.  It has 10,686 facts with 26 relations and 104 entities. From this, we created a tensor of size $104 \times 104 \times 26$. 

\textbf{UMLS} \cite{Kemp2006} has data on biomedical relationships between categorized concepts of the Unified Medical Language System. It has 6,752 facts with 49 relations and 135 entities. We created a tensor of size $135 \times 135 \times 49$.

\textbf{WN18} \cite{bordes-mlj13} contains information from WordNet \cite{miller1995wordnet}, where entities are words that  belong to \textit{synsets}, which represent sets of synonymous words.   Relations like \textit{hypernym}, \textit{holonym}, \textit{meronym} and \textit{hyponym} hold between the synsets. WN18 has 40,943 entities, 18 different relationships and more than 151,000 facts. We created a tensor of size 40,943 $\times$ 40,943 $\times$ 18.

\textbf{WN18RR} \cite{dettmers2017convolutional} is a dataset derived from WN18 that corrects some problems inherent in WN18 due to the large number of symmetric relations. These symmetric relations make it harder to create good training and testing datasets, a fact noticed by \cite{toutanova2015observed} and \cite{kadlec2017knowledge}.  For example, a training set might contain $(e_1, r_1, e_2)$ and test might contain its inverse $(e_2, r_1, e_1)$, or a fact occurring with  $e_1 and e_2$ with some relation $r_2$.

\textbf{FB13} \cite{bordes-mlj13} is a subset of a facts from Freebase \cite{bollacker2008freebase} that contains general information like ``\textit{Johnny Depp won MTV Generation Award}''. FB13 has 81,061 entities, 13 relationship and 360,517 facts. We created a tensor of size 81,061 $\times$ 81,061 $\times$ 13.

\textbf{FrameNet} \cite{baker1998berkeley} is a lexical database describing how language can be used to evoke complex representations of \texttt{Frames} describing events, relations or objects and their  participants.

For example, the \texttt{Commerce\_buy} frame represents the interrelated concepts surrounding stereotypical commercial transactions.  Frames have roles for expected participants (e.g., \texttt{Buyer}, \texttt{Goods}, \texttt{Seller}), modifiers (e.g., \texttt{Imposed\_purpose} and texttt{Period\_of\_iterations}), and inter-frame relations defining \textit{inheritance} and \textit{usage} hierarchies (e.g., \texttt{Commerce\_buy} inherits from the more general \texttt{Getting} and is inherited by the more specific \texttt{Renting}.

We processed FrameNet 1.7 to produce triples representing these frame-to-frame, frame-to-role, and frame-to-word relationships. %
FrameNet 1.7 defines roughly 1,000 frames, 10,000 lexical triggers, and 11,000 (frame-specific) roles. %
In total, we used 16 relations to describe the relationship among these items. %

\textbf{DB10k} is a real-world dataset with about 10,000 facts involving 4,397 entities of type Person (e.g., Barack Obama) and 140 relations. We used a DBpedia public SPARQL endpoint \cite{dbpediasparql} to collect the facts which were processed in the following manner. When the object value was a date or number, we replaced the object value with fixed tag. For example, \textit{``Barack Obama \quad marriedOn \quad 1992-10-03 (xsd:date)''} is processed to produce \textit{``Barack Obama \quad marriedOn \quad date"}. In case object is an entity it is left unchanged. For example ``Barack Obama \quad is-a \quad President'' as President is an entity. Such an assumption can strengthen the overall learning process as entities with similar attribute relations will tend to have similar value in the tensor. After processing, a tensor of size 4,397 $\times$ 4,397 $\times$ 140 was created.

\textbf{\fb15-237} is a dataset containing subset of the Freebase with 237 relations and nearly 15K entities. It has triples coupled textual mention obtained from ClubWeb12. More details about the dataset can be found in \cite{toutanova2015representing,toutanova2015observed}.

\subsection{Tensor creation and parameter selection}

We created a 0-1 tensor for each dataset as shown in Figure \ref{fig:slices}. If entity \textit{s} had relation \textit{r} with entity \textit{o}, then the value of (\textit{s, r, o})  entry in the tensor is set to 1, otherwise it is set to 0. Each of the created tensors was used to generate a slice-similarity matrix using Eq. \ref{eq:heuristic_function}.

\begin{table}[t]
  \centering
  \renewcommand{\arraystretch}{1.4}
  \resizebox{\textwidth}{!}{
  \begin{tabular}{|c|l|l|} \hline
    \rowcolor{gray!15}  \textbf{Hyperparameter} & \textbf{Meaning} & \textbf{Possible Values} \\ \hline
    $\lambda_{A}$ & Coefficient of the entity embedding regularizers & $\lbrace 0.0001, 0.01, 0.1, 0, 1, 10, 100, 1000 \rbrace $ \\  \hline
    $\lambda_{r}$ & Coefficient of the relation embedding regularizers & $\lbrace 0.002, 0.2, 0.01, 0.1, 0, 1, 10, 100, 1000 \rbrace$ \\ \hline
    $\lambda_{E}$ & Coefficient of the entity embedding dissimilarity penalty & $\lbrace 1, 2, 5, 10 \rbrace$ \\ \hline
    $\lambda_{sim}$ & Coefficient of the relation similarity $\mathbf{C}$ regularizer & $\lbrace 0.00002, 0.02, 0.2, 0.1, 0, 1 \rbrace$ \\ \hline
  \end{tabular}
  }
  \caption{The possible values our hyperparameters could take.}
  \label{tab:possible_hyperparams}
\end{table}

We fixed the parameters for different datasets using co-ordinate descent, changing only one hyperparameter at a time and always making a change from the best configuration of hyperparameters found so far. The number of latent variables for the compact relational tensor $\te{R}$ was set to number of relations present in the dataset. In order to capture similarity, we computed the similarity matrix \textbf{C} using various similarity metric discussed in Section \ref{slice_sim_matrix} and present results produced by \textit{Transitivity}, as it gave better performance overall. See Table \ref{tab:possible_hyperparams} for the values our hyperparameters could take.

\subsection{Evaluation protocol and metrics}

We considered \textit{\fp} as a classification task with labels \textit{correct} (i.e., value 1) for the positive class, and \textit{incorrect} (i.e., value 0) for the negative class for a given pair of entities and a relationship. We follow the same evaluation metric used in RESCAL \cite{Nickel2011}, masking the test instances during training and using area under the curve as one of the evaluation metrics.

We conducted evaluations in three different categories. The first used a \textit{stratified-uniform} sampling for which we created a stratified sampling links with 60\% \textit{correct} and 40\% \textit{incorrect}. To create the test dataset we selected ten instances from each slice for the smaller and fewer entity datasets (Kinship, UMLS, DB10k, FrameNet, and FB15-237) and 200 instances from each slice for the larger ones (WN18, FB13, and WN18RR). We masked the test instances during training. We refer to this category as uniform since all of the relation participate equally in the generated test dataset. The results from this dataset are available in Table \ref{tab:acc_macrof1_microf1_tables}

The second category used a \textit{stratified-weighted} sampling with 60\% \textit{correct} and 40\% \textit{incorrect} links, but instead of generating five test sets we used the test dataset that was publicly available and tested it on FB13 and WN18RR. The original dataset contained 5000 positive examples. We randomly sampled 60\% of these for positive instances and used the remaining 40\% to generate negative instances by replacing their objects with randomly chosen new ones. We followed a similar procedure for FB15-237. We evaluate on the datasets in Table \ref{tab:rankperformance}.

The third evaluation dataset category is \textit{balanced-weighted}. This is the dataset made publicly available by \citet{socher2013reasoning} in his Neural Tensor Network approach. For simplicity we name the dataset as FB13NTN and WN11NTN. Details of the results are explained in Section \ref{sec:comparisionntn}.

\subsection{Results and discussion of tensor based decomposition models}

In this section we provide a detailed analysis and the results of our models, which include a quantitative comparison with other tensor-based models and the impact of knowledge graph sparsity on the tensor based models. We compare our models with neural-based ones in Section \ref{sec:transE} and provide insight on how each model performs with respect to different relations. 

\subsubsection{Comparison with other tensor based models}

% float table with three subtables to a separate page 
\begin{table}[!htbp] \centering
\renewcommand{\arraystretch}{1.3}

\begin{subtable}{\textwidth} \centering
  \caption{Fact prediction performance using Area Under the Curve (AUC) as the metric}
    \begin{tabular}{|c|c|c||c|c|c|c|c|c|} \hline
\multicolumn{9}{|c|}{\textbf{Area Under the Curve}} \\ \hline
\rowcolor{gray!15} \textbf{Model Name} & \textbf{Kinship} & \textbf{UMLS} & \textbf{WN18} & \textbf{FB13} & \textbf{DB10} & \textbf{Framenet} & \textbf{WN18RR} & \textbf{\fb15-237} \\ \hline
\multicolumn{9}{|c|}{\textbf{Previous tensor factorization models}} \\ \hline
\textbf{RESCAL} & 93.24 & \textbf{88.53} & 62.13 & 65.37 & 61.27 & 82.54 & 66.63 & 92.56\\ \hline
\textbf{Non Neg RESCAL} & 92.19 & 88.37 & 83.93 & 79.13 & \textbf{81.72} & 82.6 & 68.49 & 93.03 \\ \hline
\multicolumn{9}{|c|}{\textbf{Regularized/Constrained tensor factorization models}} \\ \hline
    \textbf{Linear + Reg} & \textbf{93.99} & 88.22 & 81.86 & 80.07 & 80.79 & 78.11 & 69.15 & 90.00 \\ \hline
    \textbf{Quad + Reg} & 93.89 & 88.11 & 84.41 & 79.12 & 80.47 & 82.34 & 66.73 & \textbf{93.07} \\\hline
    \textbf{Linear + Constraint} & 92.87 & 84.71 & 80.18 & 75.79 & 80.67 & 73.64 & 66.46 & 81.88 \\\hline
    $\bigstar$ \textbf{Quad + Constraint} & 93.84 & 86.17 & \textbf{91.07} & \textbf{85.15} & 81.69 & \textbf{86.24} & \textbf{72.62} & 86.47  \\\hline
    \end{tabular}%
  \label{auc_table}%  
\end{subtable}

\ \smallskip \\

\begin{subtable}{\textwidth} \centering
  \caption{Fact prediction performance using F1 Micro as the metric}
    \begin{tabular}{|c|c|c||c|c|c|c|c|c|} \hline
    \multicolumn{9}{|c|}{\textbf{F1 Micro}} \\ \hline
\rowcolor{gray!15} \textbf{Model Name} & \textbf{Kinship} & \textbf{UMLS} & \textbf{WN18} & \textbf{FB13} & \textbf{DB10} & \textbf{Framenet} & \textbf{WN18RR} & \textbf{\fb15-237} \\ \hline
\multicolumn{9}{|c|}{\textbf{Previous tensor factorization models}} \\ \hline
\textbf{RESCAL} & 81.31 & 67.71 & 40.01 & 47.04 & 40.00  & 60.75 & 56.57 & 78.84\\\hline
\textbf{Non Negative RESCAL} & 77.23 & \textbf{69.43} & 63.69 & 58.28 &\textbf{47.73} &60.75 & 52.35 & 79.45 \\\hline
\multicolumn{9}{|c|}{\textbf{Regularized/Constrained tensor factorization models}} \\ \hline
\textbf{Linear + Reg} & \textbf{81.54} &68.04 &60.31&57.96&47.39&54.75&47.58 & 70.80 \\\hline
\textbf{Quad + Reg} & 81.38&67.35&64.09&57.22&47.39&60.62&44.92 & \textbf{79.70}\\\hline
\textbf{Linear + Constraint} & 78.46&58.73&57.04&49.15&46.13&47&46.05 & 13.70\\\hline
$\bigstar$ \textbf{Quad + Constraint} & 81.23&62.00&\textbf{79.62}&\textbf{67.88}&44.12&\textbf{66.5}&\textbf{68.01} & 59.59\\\hline
    \end{tabular}%
  \label{f1_micro}%
\end{subtable}

\ \smallskip \\

\begin{subtable}{\textwidth} \centering
  \caption{Fact prediction performance using F1 Macro as the metric}
    \begin{tabular}{|c|c|c||c|c|c|c|c|c|} \hline
    \multicolumn{9}{|c|}{\textbf{F1 Macro}} \\ \hline
\rowcolor{gray!15} \textbf{Model Name} & \textbf{Kinship} & \textbf{UMLS} & \textbf{WN18} & \textbf{FB13} & \textbf{DB10} & \textbf{Framenet} & \textbf{WN18RR} & \textbf{\fb15-237}\\ \hline
\multicolumn{9}{|c|}{\textbf{Previous tensor factorization models}} \\ \hline
\textbf{RESCAL} & 74.54&51.85&3.01&18.53&0.41&41.23&40.45 & 69.71 \\\hline
\textbf{Non Negative RESCAL} & 71.29&\textbf{55.87}&51.67&40.13&\textbf{15.06}&42.08&32.49 & 70.60\\\hline
\multicolumn{9}{|c|}{\textbf{Regularized/Constrained tensor factorization models}} \\ \hline
\textbf{Linear + Reg} & \textbf{74.77}&53.55&46.44&38.43&14.9&30.65&24.01 & 55.38\\\hline
\textbf{Quad + Reg} & 74.6&52.09&52.26&37.79&14.9&41.64&20.53 & \textbf{71.19}\\\hline
\textbf{Linear + Constraint} & 71.5&37.55&42.62&27.9&10.86&23.07&26.82 & 46.80\\\hline
$\bigstar$ \quad \textbf{Quad + Constraint} & 74.37&42.15&\textbf{78.21}&\textbf{62.41}&13.53&\textbf{58.23}&\textbf{63.5} & 36.63\\\hline
\end{tabular}
  \label{f1_macro}
\end{subtable}

% common caption
 \caption{Fact prediction performance for all models using different metrics: AUC, micro-averaged F1, and macro-averaged F1. Linear + Reg is the linear tensor decomposition with regularization on $\te{R}$. Quad + Reg, our previous work \cite{padia2016inferring}, is the quadratic tensor decomposition with regularization on $\te{R}$. Linear + Constraint and Quad + Constraint are the linear and quadratic tensor decomposition with constraints on $\te{R}$ incorporated as a Lagrange model multiplier. Transitivity similarity measure is used as prior. The $\bigstar$  next to an algorithm means it performed best overall measured with statistically significant using Wilcoxon paired rank sum test at significance level of 1\% (0.01) .}
\label{tab:acc_macrof1_microf1_tables}
\end{table}

Table \ref{tab:acc_macrof1_microf1_tables} shows the performance of all our models using three different metrics. We first focus our discussion on area under the curve (AUC), where we see our models obtain relative performance gains ranging from 5\% to 50\%. We note that AUC was the evaluation metric used by \citet{Nickel2011}, and we use it as one of our evaluation metrics for consistency. We include an in-depth examination of the different similarity encodings in Figure \ref{fig:auc-pct-change-over-rescal}, and then examine the standard information extraction F1 metric in more detail.

The Kinship and UMLS datasets have a significantly higher graph density compared to our other five datasets, as shown in Table \ref{tab:ds_stats}.  Combining this observation with the results in Table \ref{tab:acc_macrof1_microf1_tables}, we notice that graphs with lower density result in larger performance variability across both the baseline systems and our models. This suggests that when learning knowledge graph embeddings on \textit{dense} graphs, basic tensor methods with non-knowledge-graph specific regularization or constraints, such as RESCAL, could be used to give acceptable performance. On the other hand, this also suggests that for lower density graphs, different mechanisms for learning embeddings perform differently.

Focusing on the datasets with lower density graphs, we see that while the Linear+Constraint and Linear+Regularized models often matched or surpassed RESCAL, they achieved comparable or lower performance compared to their corresponding quadratic models. This is due to the fact that the distinction of the subject and object made by \textbf{A}$_1$ and \textbf{A}$_2$ embeddings tends not to hold in many of the standard datasets. That is, objects can behave as subjects (and vice versa), as is the case in WN18. Hence the distinction between the subject and the object may not always be needed. 

The performance difference between the quadratic and linear versions is high for WN18 and FB13, though the difference is relatively small for DB10k. This is largely because the DBpedia dataset includes many datatype properties, i.e., properties whose values are strings rather than entities.  In most cases the non-negative RESCAL variant outperforms the linear models. 

The Quad+Constraint model significantly outperforms RESCAL and performs relatively better compared to our other three models. This  emphasizes the importance of the flexible penalization that the Lagrange multipliers provides. Compared to RESCAL, regularization using similarity provides additional gain through the better quality of entity and relational embeddings. However, when compared to non-negative RESCAL, the regularized model performs relatively similar. We believe that for \fp, regularizing the embeddings results a similar effect as introducing high sparsity in the embeddings through non-negativity constraint. %
Compared to all others, the Quad+Constraint model performs better in most of the cases, since the Lagrange multiplier introduces flexibility in penalizing the latent relational embeddings while learning. We also conducted statistical significance using Wilcoxon rank sum paired test across all the algorithms and all datasets at significance level of 1\% (0.01) and found the Quad+Constraint model to perform better compared to the other algorithms.
    
We observe similar trends with other standard classification metrics, such as micro- or macro-averaged F1. These can be seen in Tables \ref{f1_micro} and \ref{f1_macro}, respectively. We see that, as with AUC, the Quad+Constraint model performs well overall. Meanwhile, the Linear+Reg model performs well on Kinship and comparably to the top performing system on UMLS; this reflects the prior observed connection between higher graph density and overall competitiveness of all models involved. While there can be large variability both within and across micro- and macro-F1 in the knowledge-endowed, tensor factorization models, the Quad+Constraint model yields a high performing classifier that may not be as sensitive to less-frequently occurring relations as other factorization methods. This further highlights the the knowledge encoding's positive impact.

In summary, both the Quadratic and Linear models are important depending on the data, with the Quad+Constrain model performing the best overall and the Linear models performing comparably, depending on the data.

\subsubsection{Behavior of tensor-based models and knowledge graph density}

In order to understand the behavior of different tensor based models to handle knowledge graph of different density we conducted experiments in which we reduced the number of subjects present in the graph and kept the objects constant. Reducing the number of subject with constant number of objects simulates the effect of the graph getting denser. For our experiment we used FB13 which has nearly 16K objects and 76K subjects, indicating that on an average each object entity connected to nearly five subjects. 

\begin{figure*}[h!] 
\newcommand{\myscale}{0.18}
\centering
\begin{subfigure}{.45\textwidth}
\includegraphics[scale=\myscale]{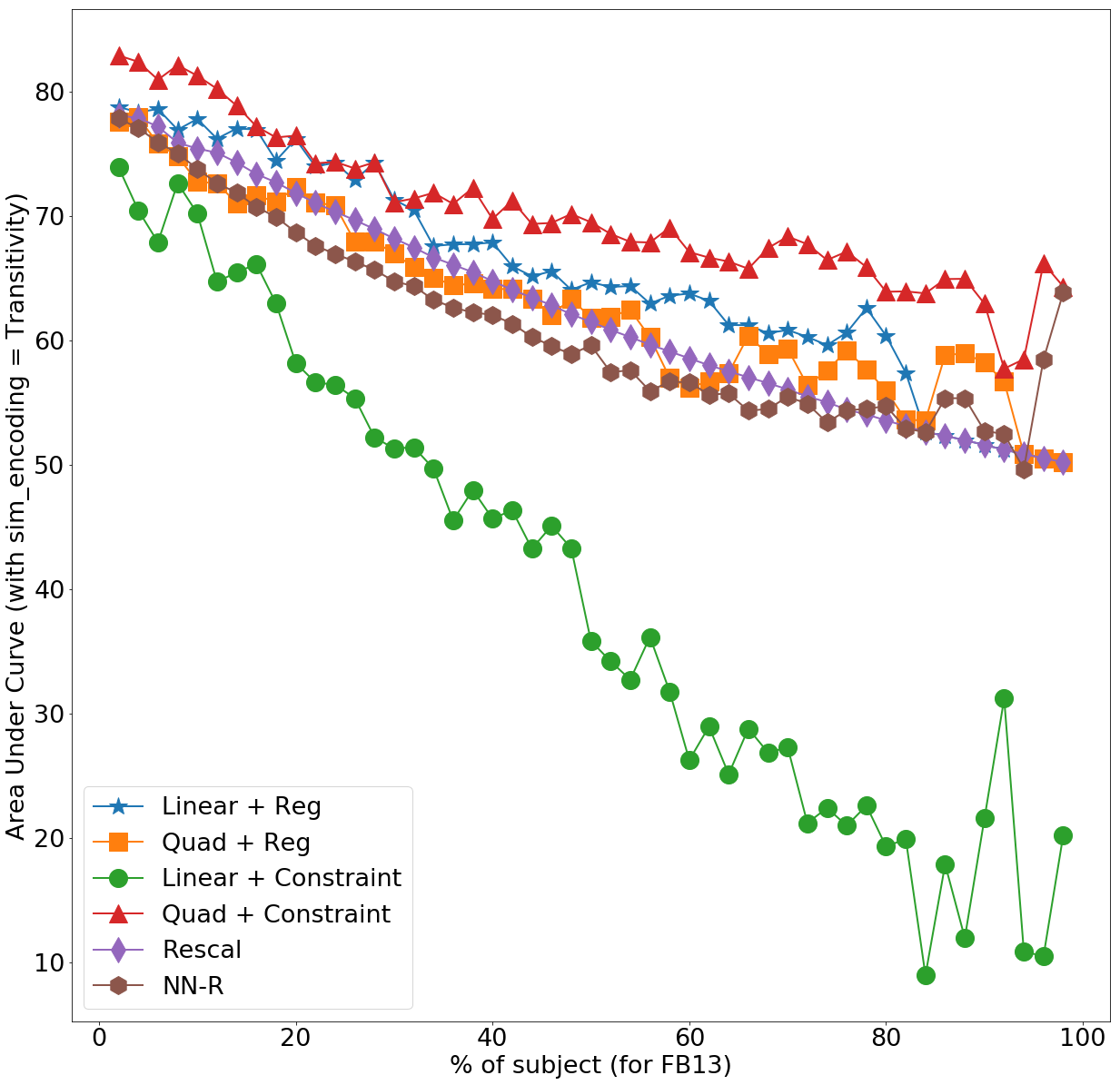}
\caption{Transitivity Similarity Matrix}
\end{subfigure}
~
\begin{subfigure}{.45\textwidth}
\includegraphics[scale=\myscale]{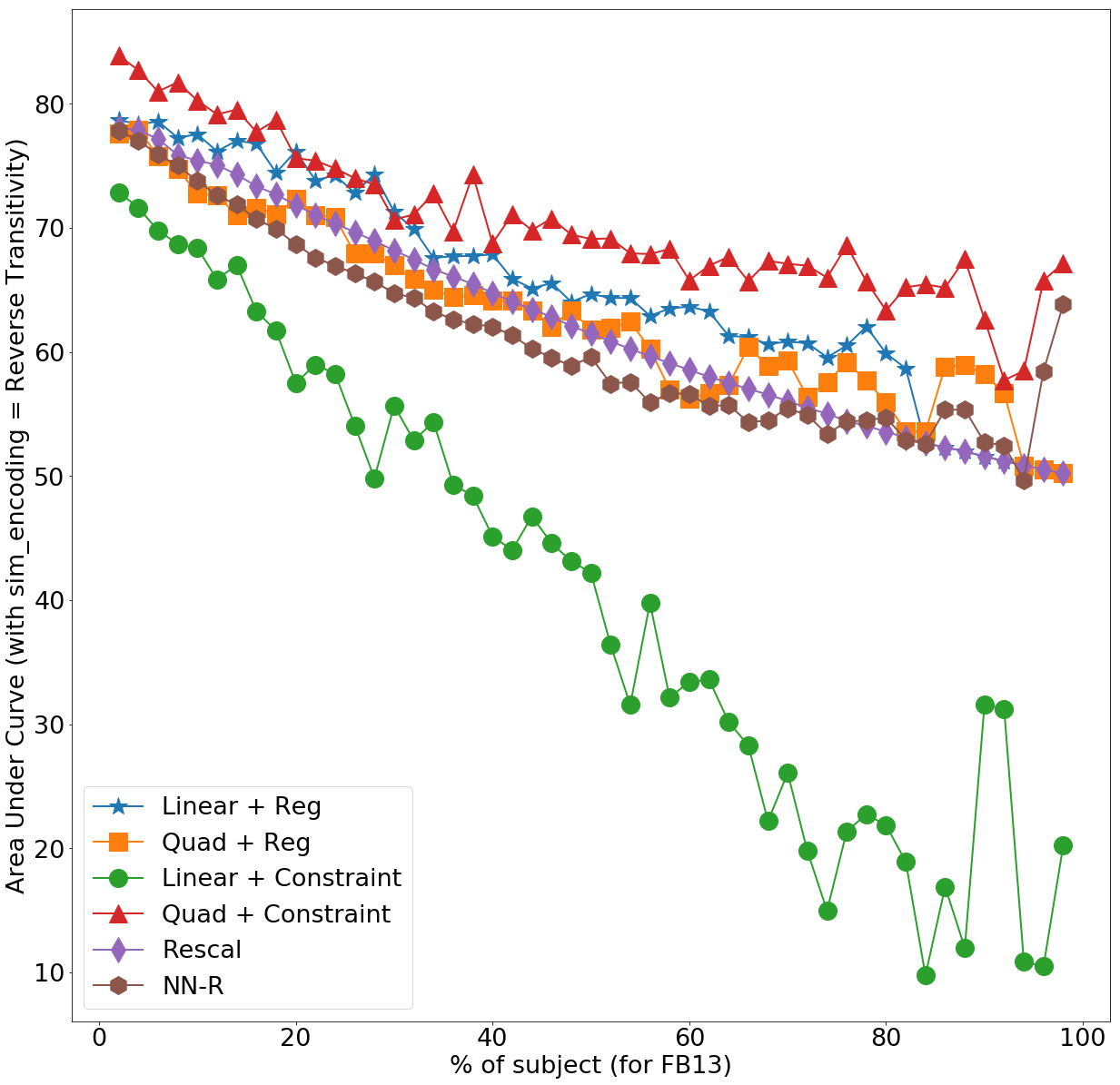}
\caption{Reverse Transitivity Similarity Matrix}
\end{subfigure}
~
\begin{subfigure}{.45\textwidth}
\includegraphics[scale=\myscale]{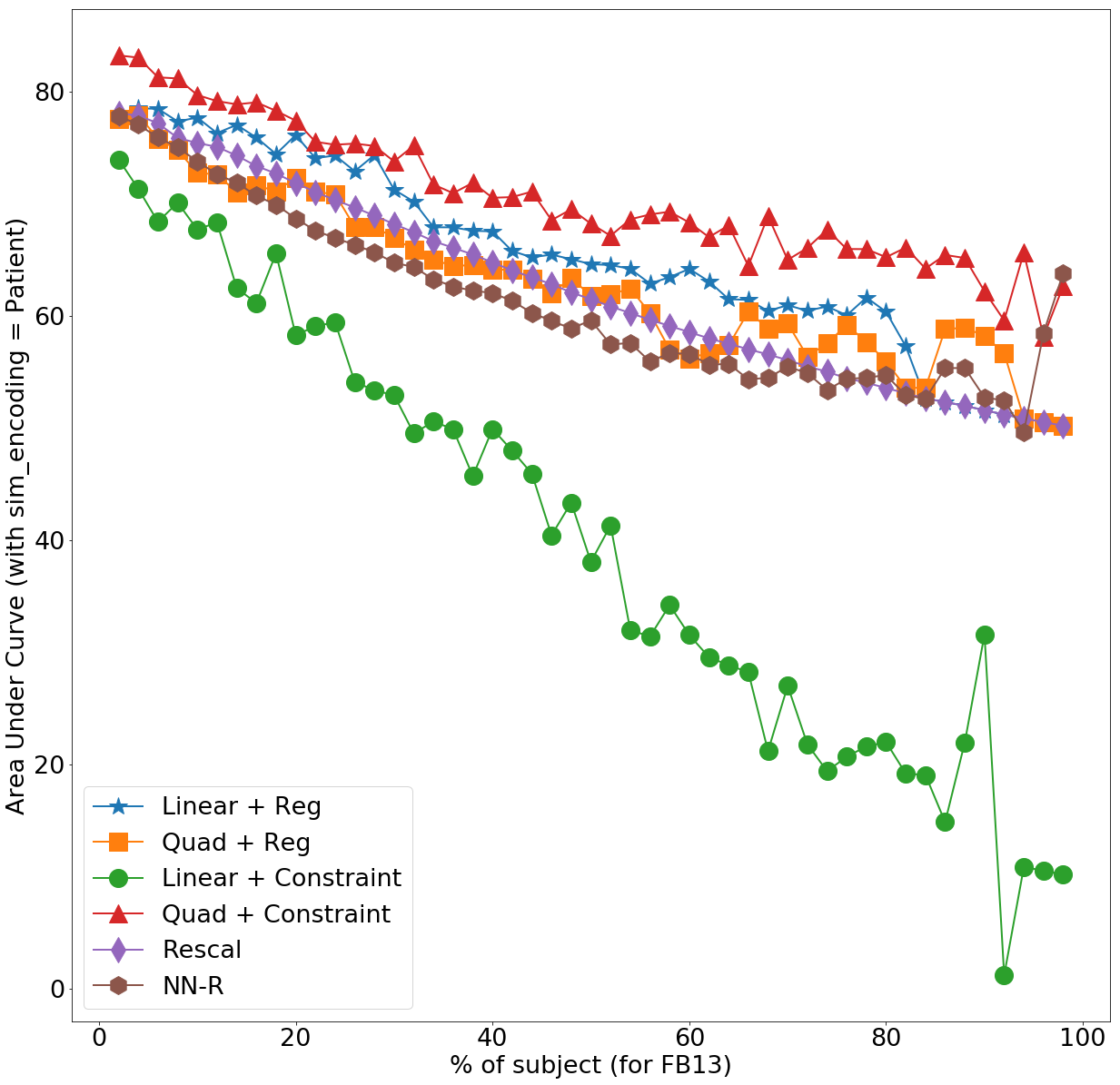}
\caption{Patient Matrix}
\end{subfigure}
~
\begin{subfigure}{.45\textwidth}
\includegraphics[scale=\myscale]{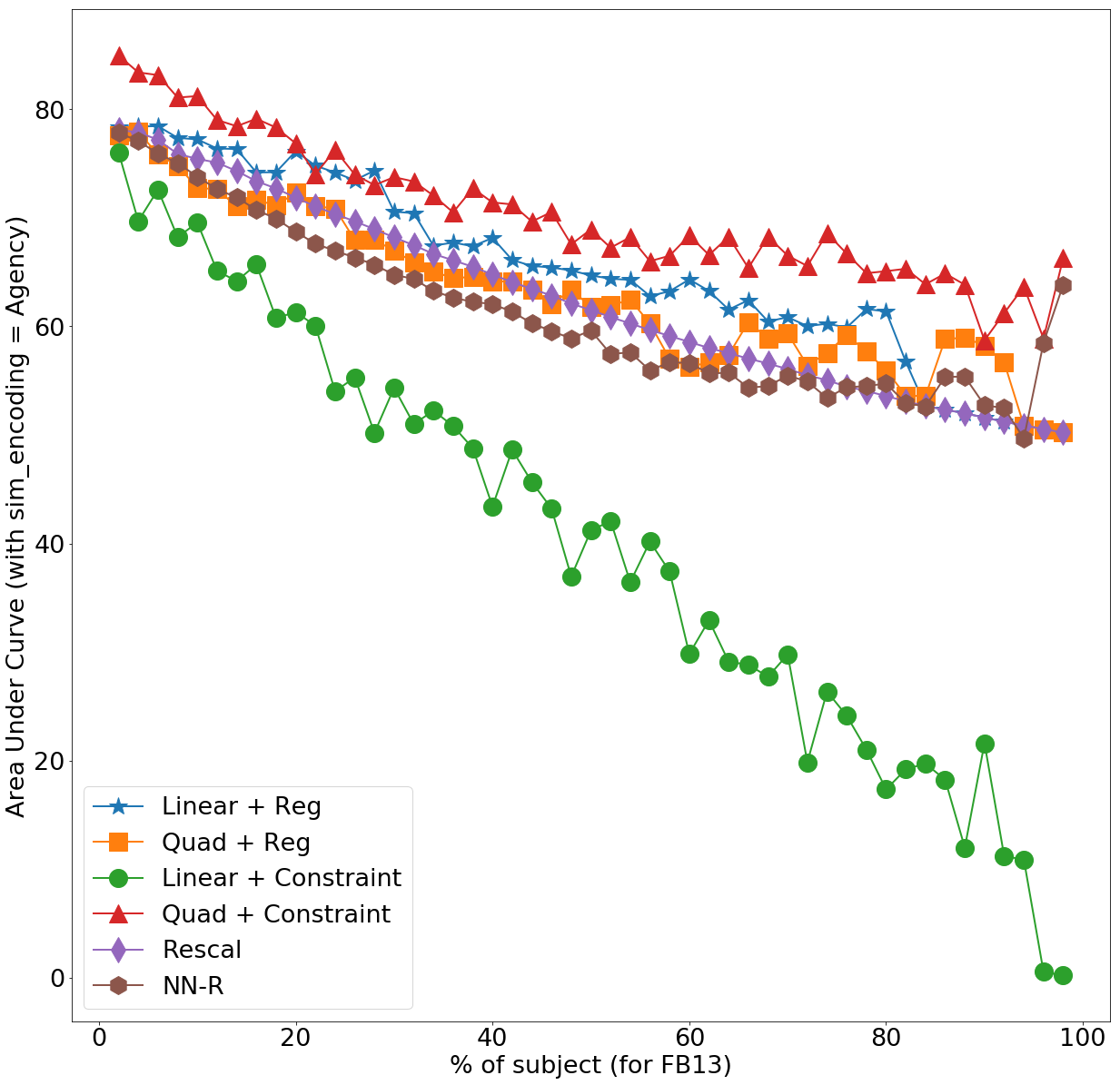}
\caption{Agency Similarity Matrix}
\end{subfigure}
\caption{Increasing performance of tensor based model when reducing \% of subjects in a knowledge graph. Here 100\% represent the original dataset.}
\label{fig:soratio}
\end{figure*}

Figure \ref{fig:soratio} shows the behavior of different tensor based models when 2\% to 100\% of the subjects are used, where 100\% represents the original dataset. Each of the tensor based models benefits when fewer subjects are considered, increasing the knowledge graph's density. Among all the models, Linear+Constraint model improves significantly faster when the number of subjects is reduced irrespective of the similarity metric, eventually achieving comparable performance with other tensor based models. The Quad+Constraint model performs the best irrespective of the graph's density.  

\subsubsection{Effect of different similarity encoding}

\begin{figure}[!t]
  \includegraphics{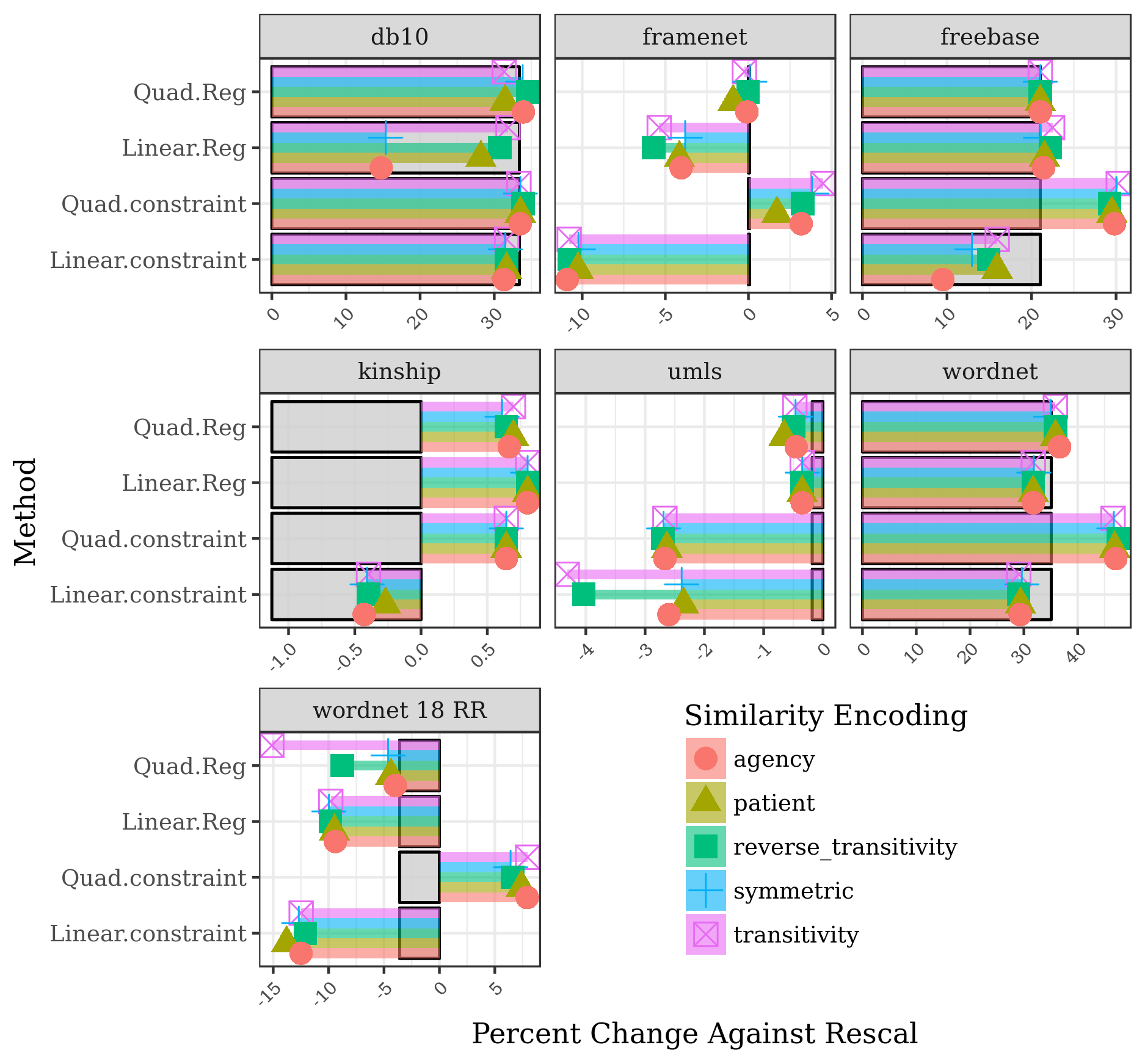}
  \caption{The percent change in AUC that our four models, each with the five different similarity encoding $C$ methods, achieve over RESCAL. The percent change that non-negative RESCAL has over RESCAL is show via the gray boxes.}
  \label{fig:auc-pct-change-over-rescal}
\end{figure}

Figure \ref{fig:auc-pct-change-over-rescal} describes the relative changes in performance of each of the similarity metrics introduced in Section \ref{slice_sim_matrix}. Here we examine the performance of these instances of our framework vs. the well studied RESCAL model. The figure shows the percent change of the methods against RESCAL (grouped by how we encode the knowledge). The gray boxes show the percent change of Non-Negative RESCAL vs. RESCAL. This shows how our approach is doing against both baselines.

Most of the similarity encoding approaches perform equally well. However, the encoding can yield a significant performance gain, especially for certain datasets. Consider the dataset db10 (top left) using Linear+Regularized. Here the \textit{agency} and \textit{symmetric} similarity encodings give poor performance. However, when the encoding is changed to \textit{transitivity} or \textit{reverse\_transitivity} there is a large gain in performance. On the other hand if WN18RR is considered, \textit{transitivity} and \textit{reverse\_transitivity} with the Linear+Regularized model both perform poorly. The Linear+Constraint model performs similarly for all kinds of encoding. Moreover, Quad+Constraint performs consistently well compared to all the baselines without being affected by the similarity encoding.

In general, while we find that different kinds of similarity encoding methods \textit{can}, and do, influence performance, on the datasets examined here we can see the effect of how that knowledge is encoded.  For example, whether a similarity encoding uses a symmetric or transitive approach may be less important than whether or not accurate knowledge is encoded at all. That is, the knowledge enrichment that the encoding provides can result in effective model generalization beyond what simple, knowledge-poor regularizations, such as a simple Frobenius norm regularization, provides.

\section{Comparison with TransE, DistMult and ComplEx}
\label{sec:transE}

This section gives a detailed comparison of our models with TransE, DistMult, and ComplEx, each of which uses a different approach to learn  embeddings of entities and relationships, as described in Section \ref{related_work}. We demonstrate that our tensor based method perform significantly better on the fact prediction task for the datasets FB13 and WN18RR and is a close second for the \fb15-237 dataset, as shown in Table \ref{tab:rankperformance}. We also show that including prior information using relation similarity results in a significant performance gain for the fact prediction task. 

\subsection{Evaluation protocol and datasets}

\textbf{Link ranking} tasks are useful for recommendation systems and have been used in previous work to determine performance of a system to predict missing links in a multi-relational data. Each fact in the data is a triple $(s,r,o)$ where $s$ and $r$ are given and each entity is treated as a potential object $o$ to predict its score and sorted rank. If the object has rank above a given threshold, it is considered a hit and is used to measure the performance of a recommendation system. 

While calculating the performance of the system, TransE considers translation from source to object for a given relation and vice-versa to calculate the mean rank. Such evaluation protocol may hold true when recommending the top-n links and may not generally hold for relations like ``hasParent'' or ``bornIn'.  For example, [Albert\_Einstein $\cdot$ bornIn $\cdot$ Germany] is a valid fact, but [Germany $\cdot$ bornIn $\cdot$ Albert\_Einstein] is not. Hence we considered translation from source to object only and compare our approach with TransE accordingly. Similarly for the DistMult and ComplEx. Moreover, as the \fp{} task is one of binary classification, we consider a fixed threshold for all relation such that if the score exceeds it, the relation is considered positive/correct else negative/incorrect.

We follow what we believe to be an advisable practice having a single threshold for all relations, rather than using hyperparameters for relation-specific thresholds that mist be tuned or learned.  Part of our motivation is knowing that the relation thresholds used in \cite{socher2013reasoning} are not publicly available. 

As TransE, DistMult, and ComplEx have been evaluated on the \lr{} task, it considers only \textit{correct} links and no \textit{incorrect} links. Hence the available dataset contains only positive examples. We consider both \textit{positive} and \textit{negative} links while comparing performance. We evaluated the performance using the standard AUC metric. We used the TransE implementation made available by the authors\footnote{\url{https://github.  com/glorotxa/SME}} and set the hyperparameters as mentioned in the paper. For DistMult and ComplEx we used the code available from the author\footnote{\url{https://github.com/ttrouill/complex}} and set the hyperparameters to find the learning rate and epoch that gave best performance.

We used the FB13, WN18RR and \fb15-237 datasets and created a training, test and validation file for each. In order to generate \textit{incorrect} links, we considered a stratified testing dataset with 60\% positive instances and randomly generated 40\% negative instances to keep testing consistent with other datasets. Negative instances were created by keeping the subject and relation fixed and randomly sampling from the pool of objects such that the result did not overlap with positive test instances. We maintained the same distribution of train, test and validation as mentioned in \cite{bordes2013translating}. As mentioned before, beside stratified-weighted sampling we considered balanced and challenged datasets available from \cite{socher2013reasoning}, which we call WN11NTN and FB13NTN, that contain equal number of positive and negative examples. We evaluated them for the sake of completeness and briefly discuss the results in the next Section.

\begin{table}[!htb]
  \centering 
  
  \caption{\normalsize Fact prediction evaluation of FB13, WN18RR and \fb15-237 by all systems and models}
  \renewcommand{\arraystretch}{1.6}
  { \footnotesize 
    \begin{tabular}{|c||c|c|c|c||c|c|c|c ||c|c|c|c|}  
    \hline
    \rowcolor{gray!15} \textbf{Dataset} & \multicolumn{4}{c||}{\textbf{FB13}} & \multicolumn{4}{c||}{\textbf{WN18RR}} & \multicolumn{4}{c|}{\textbf{\fb15-237}} \\
    \hline
    
    \rowcolor{gray!15} &  & \multicolumn{2}{c|}{\textbf{F1}} &  &  & \multicolumn{2}{c|}{\textbf{F1}}  &  & & \multicolumn{2}{c|}{\textbf{F1}} & \\ \cline{3-4}\cline{7-8}\cline{11-12}

   \rowcolor{gray!15} \multirow{-2}{*}{\textbf{Metric}} & \multirow{-2}{*}{\textbf{AUC}} & \textbf{Macro} & \textbf{Micro} & \multirow{-2}{*}{\textbf{ACC}} & \multirow{-2}{*}{\textbf{AUC}} & \textbf{Macro} & \textbf{Micro} & \multirow{-2}{*}{\textbf{ACC}} & \multirow{-2}{*}{\textbf{AUC}} & \textbf{Macro} & \textbf{Micro} & \multirow{-2}{*}{\textbf{ACC}} \\
    \hline
    \textbf{Rescal } & 80    & 4.08  & 40.00    & 40.00    & 69.63 & 38.18 & 49.8  & 49.8 & 97.61 &  94.03 & 73.41 & 94.03 \\
    \hline
    \textbf{Non Neg Rescal} & 77.76 & 40.95 & 51.38 & 51.38 & 67.41 & 67.41 & 45.52 & 45.52 & \textbf{97.81}  & \textbf{94.48} & 73.59 & \textbf{94.48} \\\hline
    \hline
    \textbf{TransE} & 52.3  & 16.497 & 40.72 & 42.76 & 68.39 & 46.91 & 62.1  & 62.08 & 50.84  & 41.11 & 4.21 & 41.12  \\
    \hline
    \textbf{DistMult} & 54.36 & 29.2  & 53.76 & 61.97 & 67.39 & 37.76 & 61 & 60.88  & 70.28  & 64.65 & 45.33 & 64.65 \\
    \hline
    \textbf{ComplEx} & 61.09 & 28.86 & 54.06 & 53.32 & 67.61 & 34.235 & 60.88 & 61.14 & 67.64 & 61.59 & 35.12 & 61.59 \\\hline
    \hline
    \textbf{Linear+Reg} & 76.51 & 35.44 & 50.6  & 55.08 & 68.69 & 28.95 & 48.9  & 48.9  & 96.49 & 91.08  & 59.59 & 91.08 \\
    \hline
    \textbf{Quad+Reg} & 75.15 & 34.85 & 50.52 & 54.62 & 68.46 & 17.96 & 48.5  & 48.5 & 97.2    & 92.91 & \textbf{73.8} & 92.91\\
    \hline
    \textbf{Linear+Constraint} & 73.23 & 27.82 & 44.72 & 47.04 & 66.66 & 26.99 & 44.72 & 44.72  & 80.00  & 43.53 & 1.06 & 43.53\\
    \hline
    \textbf{Quad+Constraint} & \textbf{82.49} & \textbf{56.48} & \textbf{59.04} & \textbf{66.48} & \textbf{81.86} & \textbf{59.09} & \textbf{62.54} & \textbf{65.49} & 94.59 & 84.34 & 53.56 	 & 84.34 \\
    \hline
    \end{tabular} }%
  \label{tab:rankperformance}%
\end{table}%

\subsection{Analysis and discussion}

\label{sec:comparisionntn}
Table \ref{tab:rankperformance} shows the performance of previous tensor based models, with TransE, DistMult, ComplEx and our models. Our Quad+Constraint model provides significant improvement over TransE, DistMult, and ComplEx. 

One reason our models outperform TransE and DistMult is that the embeddings learned by these system is task-specific and are more suitable for a \lr{} task than for a \fp{} one. The results for ComplEx suggest that the embedding learning method in complex space is work better for \lr{} than \fp{}. When comparing the baseline methods DistMult and ComplEx on balanced WN11NTN and FB13NTN datasets, our approach performed better with a 4-5\% absolute improvement, indicating that the current embedding based method are better suited for \lr than \fp.

For the FB15-237 dataset with 237 relations, the quadratic based tensor models, i.e., Rescal, Non-Negative Rescal, Quad+Reg, and Quad+Constraint, give comparable or best AUC scores compared to TransE, DistMult, and ComplEx. Moreover, considering other metrics, the quadratic models, either regularized or constrained, perform better overall (as seen in the F1-Macro performance) and also at individual level (as seen in F1-Micro). On the other hand,  the lower score of the Linear+Constraint model is due to it frequently predicting a given fact to be incorrect. Comparing the performance of linear models, we note that regularizing embedding model performs better than the constraint one and that the quadratic versions dominate their linear counterparts. 

A review of the results in Tables 4 and 5 show that our Quad+Constraint model is better overall, and there is significance improvement when graph density is very low. We believe that the lack of information inherent in a relatively sparse graph is better captured by the constraint introduced by the similarity term. Moreover, Table 5 suggests that the embedding learned using DistMult, ComplEx and TransE work well for a link ranking task and less so for a fact prediction one. In contrast the tensor based model perform better at fact prediction task. Moreover, RESCAL and Non-Negative RESCAL perform poorly compared to our models when the graph density is low, which again demonstrate the effect of constraining the embedding using similarity for a fact prediction task.

\subsection{Per Relation Analysis with F1-macro and F1-micro}

\begin{figure}
  \centering
  \includegraphics{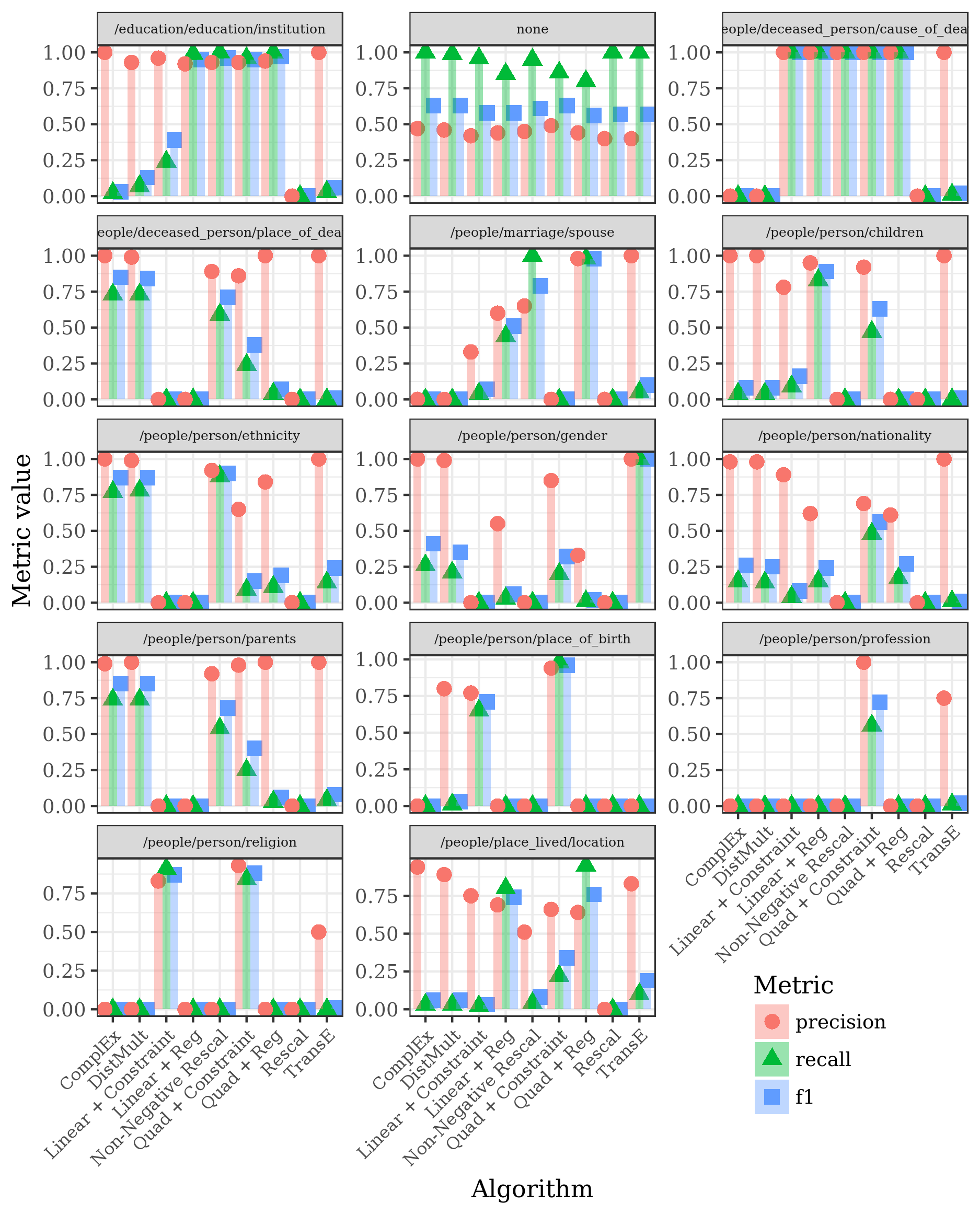}
  \caption{Precision, recall, and F$_1$ per relation for the FB13 dataset.}
  \label{fig:fb13_per_relation}
\end{figure}
\begin{figure}
  \centering
  \includegraphics{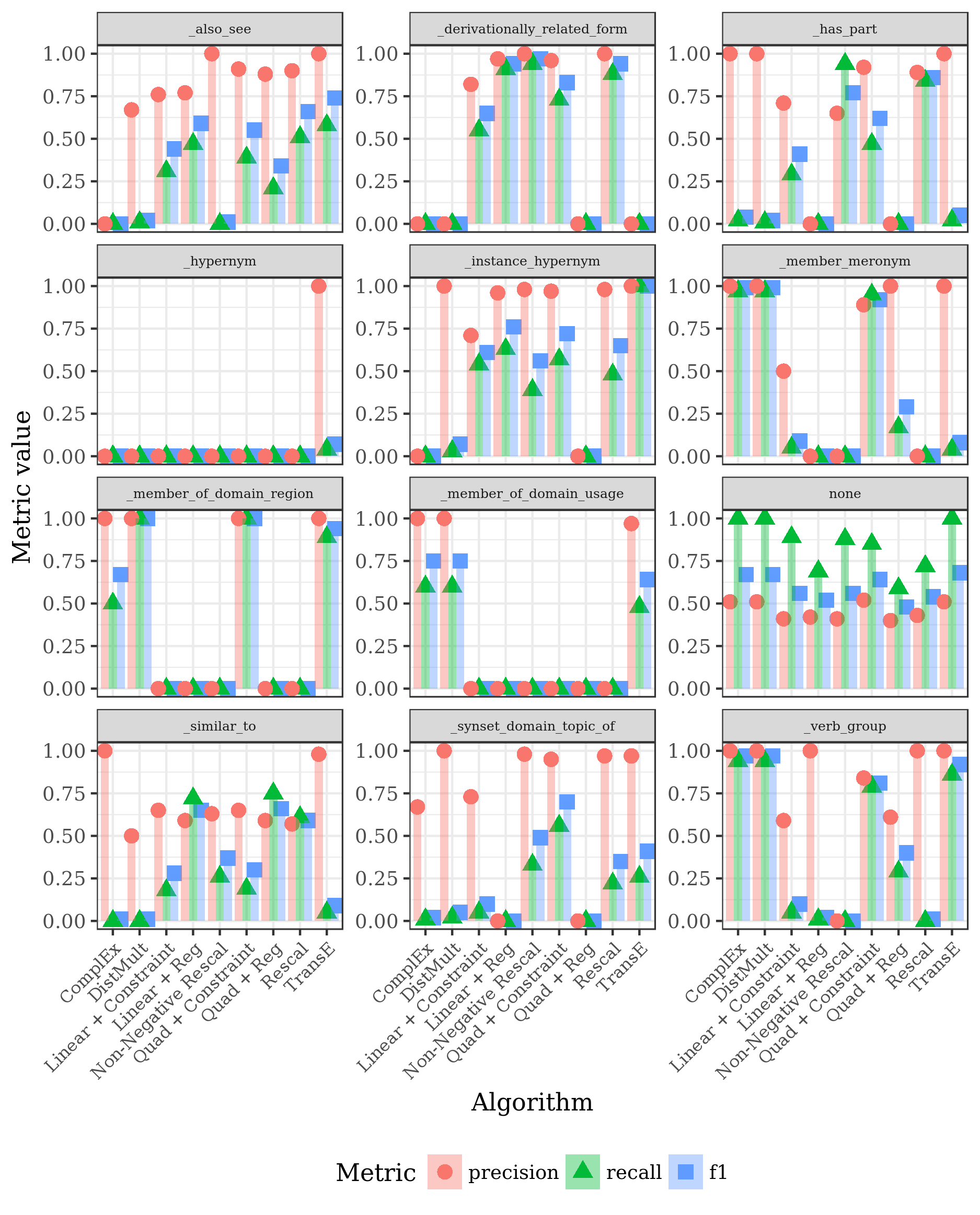}
  \caption{Precision, recall, and F$_1$ per relation for the WN18RR dataset.}
  \label{fig:wn18rr_per_relation}
\end{figure}

Figures \ref{fig:fb13_per_relation} and \ref{fig:wn18rr_per_relation} show the models' performance broken down by relation.  For simplicity of analysis, we selected the WN18RR and FB13 datasets. 
We first consider FB13. To better understand these results we argue we can group the FB13 relations in to three categories: (i) logically symmetric relations, (ii) knowledge-graph transitive relations, and (iii) what we refer to as \textit{hub} relations. 

Logically symmetric relations, like {people/marriage/spouse}, satisfy the normal definition of a symmetric relation: for a relation $r$ and entities $x$ and $y$, if $r(x,y)$ is true, then $r(y,x)$ is also true. We identified only one FB13 logically symmetric relation. This contrasts with KG transitive relations that, while not necessarily representing logically transitive relations, can be productively combined with other relations to form meaningful relation chains. FB13 KG transitive relations are {people/person/children}, {people/place\_lived/location}, {people/person/parents}. Finally, we identify the remaining FB13 relations as \textit{hub} relations that do not readily yield logically symmetric nor knowledge graph transitive relations. For example, subjects of hub relations like {people/deceased\_person/cause\_of\_death} and {person/person/nationality} cannot be easily used, under the FB13 schema, as objects of other hub relations.\footnote{We identify the following FB13 relations as hub relations: {people/deceased\_person/cause\_of\_death}, {person/person/nationality}, {people/person/place\_of\_birth}, {education/education/institution}, {people/person/gender},  {people/person/place\_of\_death}, {people/person/religion}, {people/person/ethnicity}, and {people/person/profession}.}\

Both ComplEx and DistMult generally have high precision but suffer from low recall resulting in poor F1 scores. On the other hand, TransE gives high precision for hub relations. For logically symmetric relations like {spouse}, Quad+Regularized does well, which makes sense as the relationship is two-way and is captured by the quadratic objective function. Moreover, Linear+Constraint performs poorly as it tries to model the behavior in opposition to the reality that either of the relation's arguments could be used as the subject or the object. Quad+Constraint performs better compared to other models across all relation except {spouse},  indicating that such symmetric relations are better modeled with regularization than a Lagrangian constraint. 

Second, we examine the relation-level F1 performance on WN18RR. As seen in Figure \ref{fig:wn18rr_per_relation}, the Quad+Constraint model performs consistently well across all of the relations---especially when compared to the other methods. We believe that this stems from the way similarity is incorporated and the embeddings learned. For example, consider a relation {\_synset\_domain\_topic\_of} and the heatmap shown in Figure \ref{fig:heatmap:wn18}. We believe the better performance stems from the level of similarity shared between the four relations---{\_synset\_domain\_topic\_of}, {\_instance\_hypernym}, {\_derivationally\_relation\_form}, and {\_has\_part}.

\section{Time complexity}
\label{sec:timecomplexity}

\begin{table}[t]
\centering
\renewcommand{\arraystretch}{1.4}
\caption{Running times (in seconds) per iteration  for each of the algorithms on the eight evaluation datasets; -- means not available.}%
\begin{tabular}{|c|c|c|c|c|c|c|c|c|} \hline
\rowcolor{gray!15} & \textbf{Kinship} & \textbf{UMLS} & \textbf{WN18} & \textbf{FB13} & \textbf{DB10k} & \textbf{FrameNet} & \textbf{WN18RR} & \textbf{FB15-237} \\
\rowcolor{gray!15} {\em relations} & \textit{26} & \textit{49} & \textit{18} & \textit{13} & \textit{140} & \textit{16} & \textit{11} & \textit{237}\\
\rowcolor{gray!15} {\em entities} & \textit{104} & \textit{135} & \textit{40,943} & \textit{81,061} & \textit{4,397} & \textit{62,344} & \textit{40,943} & \textit{14,541}\\
\hline
\textbf{RESCAL} 		& 0.01 	& 0.07 	& 0.26 & 0.29 	& 1.7 & 0.1 & 0.11 & 21.96\\
\textbf{NN-R} 			& 0.01	& 0.08 	& 0.38 & 0.41 	& 8.72 & 0.22 & 0.12 & 46.31\\
\textbf{TransE} 		& -- 	& -- 	& 8.26 & 19.37 	& -- & --  & 4.45 & --\\ 
\textbf{DistMult} 		& -- 	& -- 	& 20.27 & 56.7 	& -- & --  & 14.4 & 48.93 \\ 
\textbf{ComplEx} 		& -- 	& -- 	& 71.86 & 156.24 	& -- & -- & 41.3  & 157.53 \\ \hline
\textbf{Linear+Regularized} 		& 0.02 	& 0.04	& 0.85 & 1.02 	& 4.86 & 0.39 & 0.43 & 23.38 \\ 
\textbf{Quad+Regularized} 		& 0.03	& 0.11	& 0.52 & 0.62 	& 4.78 & 0.24 & 0.25 & 52.96 \\ 
\textbf{Linear+Constrained} 		& 0.02 	& 0.1 	& 0.33 & 0.39 	& 3.13 & 0.15 & 0.15 & 43.68 \\ 
\textbf{Quad+Constrained} 		& 0.03 	& 0.09 	& 0.69 & 0.71 	& 3.26 & 0.12& 0.12 & 54.28\\ 
\hline
\end{tabular}%
\label{tab:timeComplexity}%
\end{table}

The asymptotic time and space complexity of our models are the same as RESCAL's. Table \ref{tab:timeComplexity} shows the run times per iteration taken by each approach to update the unknown variables. We ran all for a maximum of 100 iterations and report the average running time per iteration. In the case of TransE, we considered an epoch as an iteration, since each iteration sees all the data values.

As expected, the running time increases with the number of relations, with the \fb15-237 dataset, which has 237 relations, taking the longest and DB10K with 140 relation in second place. The non-negative constraint on non-negative RESCAL increases the running time of that model. Regularized models require more time when compared to other models due to presence of additional terms to bring \textbf{A}$_1$ and \textbf{A}$_2$ closer, which introduces additional computation during the update rules. 

\begin{figure}[tp] %forces figure onto page by itself	
  \newcommand{\myscale}{0.280}
  \centering
  %\begin{subfigure}{.48\textwidth}
  \begin{subfigure}{\textwidth}
  \includegraphics[scale=\myscale]{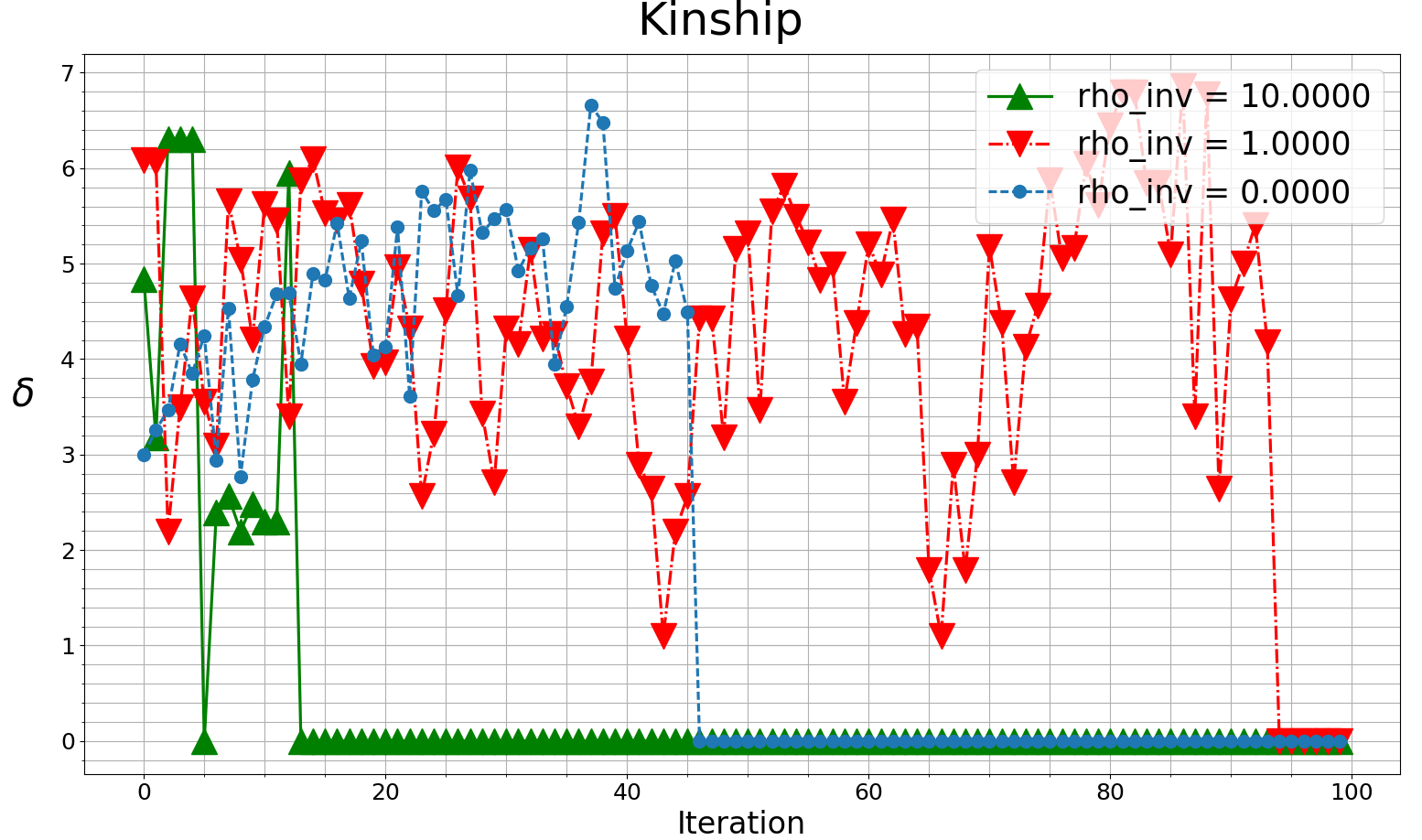}
  \caption{Convergence on Kinship}
  \end{subfigure}
 % ~

  \ \bigskip \\
  %\begin{subfigure}{.48\textwidth}
  \begin{subfigure}{\textwidth}
  \includegraphics[scale=\myscale]{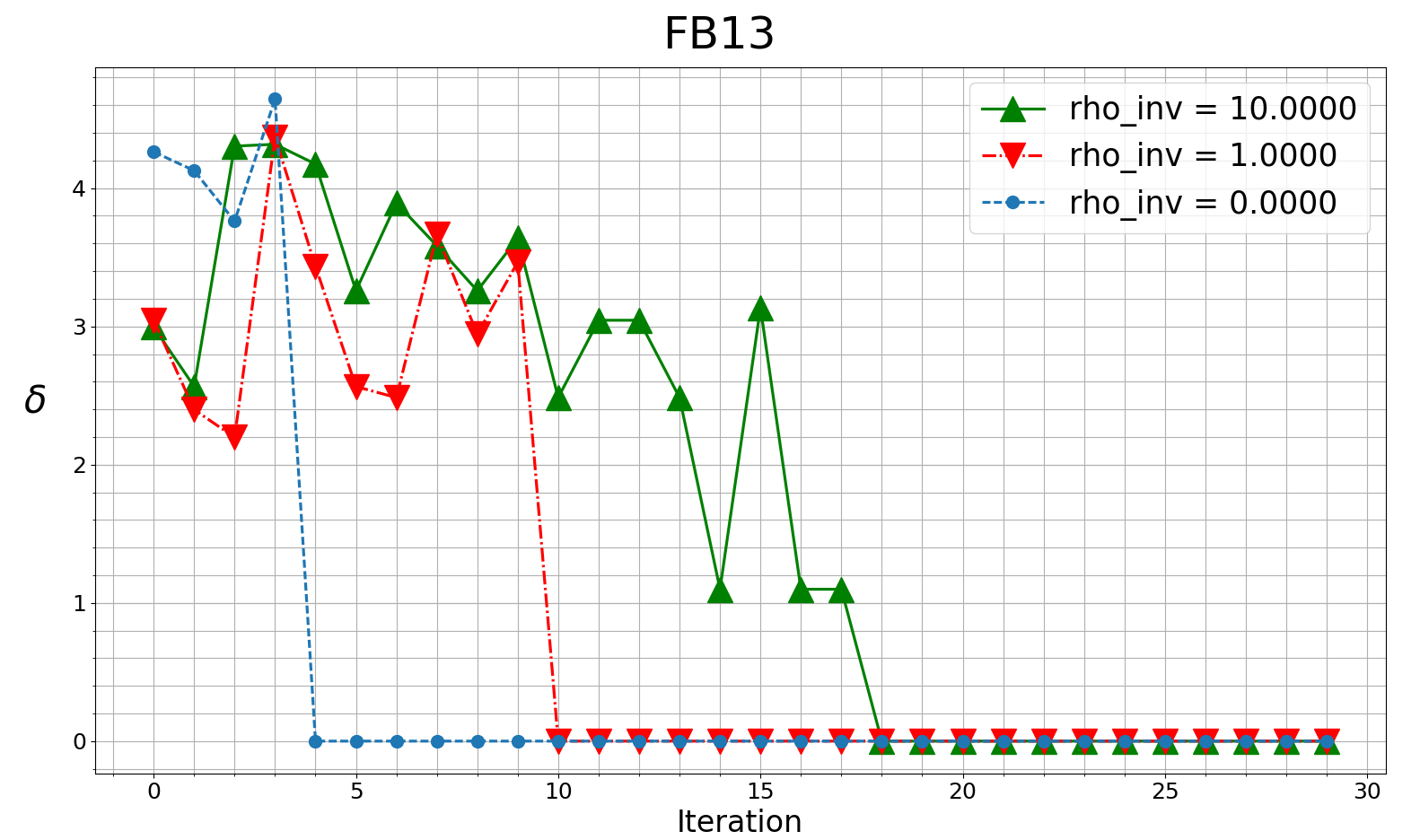}
  \caption{Convergence on FB13}
  \end{subfigure}
  \caption{Change in the unknowns/variables at each iteration for three different values of $\rho=0.1, 1, \infty$. Here \textit{rho\_inv} equals of $1/\rho$.}
  \label{fig:convergence_plot}        
\end{figure}

%% \begin{figure}[tp] %forces figure onto page by itself	
%%   \newcommand{\myscale}{0.25}
%%   \centering
%%   \subfigure[Convergence on Kinship.]{\includegraphics[scale=\myscale]{./gray_image/kinship_rho.png}}
%%   \subfigure[Convergence on FB13.]{\includegraphics[scale=\myscale]{./gray_image/freebase_rho.png}}
%%   \caption{Change in the unknowns/variables at each iteration for three different values of $\rho=0.1, 1, \infty$. Here \textit{rho\_inv} equals of $1/\rho$.}
%%   \label{fig:convergence_plot}        
%% \end{figure}

TransE has much longer running time per iteration since it computes pairwise distances among positive and negative instances, so its time increases with the number of entities. Similarly, DistMult and ComplEx consume considerable amount of time as we believe it is due to running on CPU. We also saw that running on a GPU, DistMult executed faster than TransE (3.73sec for WN18, 10.33sec for FB13, and 2.75sec for WN18RR) and ComplEx took longer than TransE (9.77sec for WN18, 21.2sec for FB13, and 8.75sec for WN18RR).

\section{Effect of \texorpdfstring{$\rho$}{rho} on convergence}
\label{sec:rho}

To illustrate convergence of the Linear+Regularized model, Figure \ref{fig:convergence_plot} shows the effect of $\rho$ on the maximum component--wise relative change in consecutive iterations for the variables during optimization. If $\boldsymbol{z}_t$ is the vector of all of the unknowns at iteration $t$, i.e. \textbf{A}$_1$, \textbf{A}$_2$, $\te{R}$, we use the following equation to measure the maximum relative change on the unknowns at each iteration.
\begin{eqnarray}
\delta(\boldsymbol{z}_t, \boldsymbol{z}_{t+1}) & = & \max_{i} \left|
\frac{z_{t}(i)-z_{t+1}(i)}
{z_{t}(i)+z_{t+1}(i)/2}
\right|
\label{eq:deltaFunction}
\end{eqnarray}
For each value of $\rho$ we follow a cold-start procedure, i.e., for each new value of $\rho$ we randomly initialize all the variables. The termination condition is that we reached the maximum iteration number (chosen as 100) or that the maximum change $\delta $ in the unknown is below a threshold (chosen as 10$^{-6}$). Here, the blue dots with dashed lines indicate the maximum relative change $\delta$ vs. iterations  when $\rho = \infty$.

\section{Conclusions and future work}
\label{conclusion}

We proposed a framework for learning knowledge-endowed entity and relation embeddings. The framework includes four readily obtainable novel models that generalize existing efforts. Two of the models optimize a linear factorization objective and two a quadratic one.  We evaluated the quality of embeddings on the task of \fp  and demonstrated  significant improvements ranging from 5\% to 50\% over state-of-the-art tensor decomposition models and translation based models on a number of real-world datasets. We motivated and empirically explored different methods for encoding prior knowledge into the tensor factorization algorithm, finding that using transitive relationship chains resulted in the highest overall performance among our models.

We observed that for the task of fact prediction, better embeddings are obtained by the Quadratic+Constrained model. Linear models are better suited when there is a one way interaction from subject to the object in which the object cannot also serve as a subject. We find the quadratic models perform better in general, irrespective of the position of the entity as subject or object. Constraint-based models perform better compared to regularized models and constraint-based models with a quadratic objective are better suited for the task---irrespective of the sparsity of the knowledge graph. We showed detailed experimental analyses of the model's strengths and weaknesses in predicting facts with particular relations, and we provided complementary qualitative analysis of commonalities among those relations. On the theoretical side, we proved that the Linear+Regularized model has the desirable property of convergence and illustrated its convergence on two standard benchmark datasets. 

Our future work will explore the use of our models in several application contexts that use natural language understanding systems to extract entities, relations and events from text documents. Such systems can benefit from a \textit{fact prediction} module that can help eliminate facts extracted in error. Our experience in the NIST Knowledge Base Population (KBP) tasks \cite{finin2015} showed the need to independently assess the quality of extracted relations.  The KBP tasks are well suited for an approach like our trained on general-purpose knowledge graphs like DBpedia, Freebase and Wikidata.

A second application is a system we are developing to identify possible cybersecurity attacks from data collected from host computers and networks represented in an RDF knowledge graph using the Unified Cybersecurity Ontology \cite{UCO16}.  This system \cite{Narayanan18} draws on background knowledge encoded in graphs populated with information extracted from cybersecurity-related documents and from semi-structured data from cybersecurity threat intelligence data sources.

A third application is as one component of a general-purpose system under development for cleaning noisy knowledge graphs \cite{Cleaning_Noisy_Knowledge_Graphs}.  Its current architecture consists of an ensemble of modules that try to identify, characterize and explain different types of errors that many current text information extraction systems can make.  Using a version of our approach to see of an extracted fact is predicted or not will be a useful feature.

\vspace{1em}
\noindent \textbf{Acknowledgement}
Partial support for this research was provided by gifts from IBM through the IBM AI Horizons Network and from Northop Grumman Corporation.

\section*{References}
\bibliography{mybibfile}

\appendix
\section{Proof of convergence}
\label{sec:appendix}

\subsection{Propositions and Lemma}
\label{sec:prepositionAndLemma}

We assume that order--3 tensors 
$\tensor{X} \in \reals^{N \times N \times K}$, and $\tensor{R} \in \reals^{N' \times N' \times K'}$, matrix 
$\bm{A} \in \reals^{N \times N'}$, and symmetric matrix $\bm{C} = (c_{ij}) \in \reals^{K \times K}$.

\begin{proposition} 
	For any matrices $\bm{A}, \bm{B}, \bm{C}, \bm{D}$ we use the following properties of Kronecker products:
	\begin{eqnarray}
	( \bm{A} \otimes \bm{B} ) ( \bm{C} \otimes \bm{D}) = (\bm{A} \bm{C} ) \otimes (\bm{B} \bm{D}) \\
	( \bm{A} \otimes \bm{B} )^T = \bm{A}^T \otimes \bm{B}^T \\
	(\bm{A} \otimes \bm{B} ) ^{1/2} = \bm{A}^{1/2} \otimes \bm{B}^{1/2} \\
	\pinv{( \bm{A} \otimes \bm{B} )} = \pinv{\bm{A}} \otimes \pinv{\bm{B}} \\
	( \bm{A} \otimes \bm{I}_n + \alpha \bm{I}_m)^{-1} = (\bm{A} + \alpha\bm{I}_{m-n})^{-1} \otimes \bm{I}_n \\
	\bm{A} \otimes (\bm{B} \otimes \bm{C} ) = ( \bm{A} \otimes \bm{B} ) \otimes \bm{C} \\
	\tvec{ \bm{A} \bm{B} \bm{C} } = ( \bm{C}^T \otimes \bm{A} ) \tvec{\bm{B}}
	\end{eqnarray}
\end{proposition}
\begin{proposition}[\cite{Kolda2009}]
	For order--N tensors $\tensor{X}, \tensor{Y}$ and sequence of matrices $\mseq{A}{i}$, $i=1,2,\ldots,N$, 
	\bgeq
	\tensor{Y} =  
	\tensor{X} \tmmul{1} \mseq{A}{1} \tmmul{2} \mseq{A}{2} \ldots \tmmul{N} \mseq{A}{N}  
	\eneq
	if and only if, for all $n$
	\bgeq
	\tfold{Y}{n} =   \mseq{A}{n} \tfold{X}{n} 
	\left( \mseq{A}{N} \otimes \ldots \otimes \mseq{A}{n+1} \otimes \mseq{A}{n-1} \ldots 
	\otimes \mseq{A}{1} \right)^T 
	\eneq
	Further, 
	\bgeq
	\tensor{Y} = \tensor{X} \tmmul{n} \bm{A}  \ \mbox{iff}\ \tfold{Y}{n} = \bm{A} \tfold{X}{n} 
	\eneq
	\bgeq
	\fnorm{\tensor{X}}^2 = \sum_{i} \fnorm{\fslice{X}{i}}^2 = \sum_{i} \norm{\tvec{\bm{X}_i}}^2_2 = \norm{\tvec{\tensor{X}}}^2_2
	\eneq
\end{proposition}
\begin{proposition}
	For any order--3 tensors $\tensor{X}$ and $\tensor{Y} = \tensor{X} \tmmul{1} \bm{A}_1 \tmmul{2} \bm{A}_2$, we have
	the following. Both mode-1 and mode-2 unfoldings of order--3 tensors are block matrices with the same number of blocks, and 
	\begin{eqnarray}
	\tfold{X}{2} & = & [ \fslice{X}{1}' \ldots \fslice{X}{k}' \ldots], \\ 
	\tfold{Y}{1} & = & \bm{A}_1 \tfold{X}{1} (\bm{I} \otimes \bm{A}_2)^T= [  \ldots \bm{A}_1 \fslice{X}{k} \bm{A}^T_2 \ldots ] \\
	\tfold{Y}{2} & = & \bm{A}_2 \tfold{X}{2} (\bm{I} \otimes \bm{A}_1)^T= [  \ldots \bm{A}_2 \fslice{X}{k}' \bm{A}^T_1 \ldots ]
	\end{eqnarray}
\end{proposition}
\begin{proposition}
	\begin{eqnarray}
	\hspace*{-3.5cm}\argmin_{\bm{X}} \fnorm{ \bm{B} - \bm{A} \bm{X} \bm{C} }^2 = 
	\argmin_{\bm{X}} \norm{ \tvec{\bm{B}} - (\bm{C}^T \otimes \bm{A} ) \tvec{\bm{X}}}^2
	\end{eqnarray}
	and the solution of a least squares problem 
	\begin{eqnarray}
	\bm{x} & \leftarrow & \argmin_{\bm{x}} \norm{ \bm{A} \bm{x} - \bm{b}}^2 \\ 
	&  = & \argmin_{\bm{x}} ( \bm{x}^T (\bm{A}^T \bm{A} ) \bm{x} - 2 (\bm{A}^T \bm{b})^T \bm{x} + \bm{b}^T \bm{b} )= \\
	& = & (\bm{A}^T \bm{A})^{-1} \bm{A}^T \bm{b} = 
	\pinv{\bm{A}} \bm{b} 
	\end{eqnarray}
	where $\pinv{\bm{A}}$ the Moore--Penrose matrix pseudo--inverse 
	(provided $\bm{A}^T \bm{A}$ is full--rank and hence invertible), 
	which is a left--inverse of $\bm{A}$ with least Frobenius norm among all left--inverses of $\bm{A}$.
	Furthermore, its gradient and Hessian are
	$2 \bm{A}^T \bm{A} \bm{x} - 2 \bm{A}^T \bm{b}$
	and $2 \bm{A}^T \bm{A}$ respectively.
\end{proposition}
\begin{lemma}
	\label{lemma:sim3}
	For any order--3 tensor $\tensor{X}$ and symmetric matrix $\bm{C}$, we have that
	\bgeq
	\fnorm{ \tensor{X} \tmmul{3} (\mathrm{diag}(\bm{C} \cdot \bm{1}_{n_2} ) -\bm{C})^{1/2} }^2 = 
	\sum_{ij} c_{ij} \fnorm{\fslice{X}{i} - \fslice{X}{j}}^2 
	\eneq
\end{lemma}
\proof
Suppose that $\tensor{X} \in \reals^{n_1 \times n_2 \times n_3}$.
Consider the tensor's mode--1 unfolding $\tfold{X}{1} = [\fslice{X}{1} \fslice{X}{2} \ldots \fslice{X}{K} ]$.
Clearly, $\tvec{\tfold{X}{1}} = \tvec{[\tvec{\fslice{X}{1}} \ldots  \tvec{\fslice{X}{K}}]}$.
If $\bm{L} = \mrm{deg}(\bm{C}) - \bm{C}$, where $\mrm{deg}(\bm{C})$ 
is a diagonal matrix with the row sums of $\bm{C}$, 
then recall ($\bm{L}$ is the Laplacian matrix of an undirected graph with weighted adjacency matrix $\bm{C}$) that 
\bgeq
\bm{x}^T \bm{L} \bm{x} = \sum_{ij} c_{ij} (x_i - x_j)^2.
\eneq
Consequently, we have
\begin{eqnarray}
\sum_{ij} c_{ij} \fnorm{\fslice{X}{i} - \fslice{X}{j}}^2  &=&  
\tvec{\tfold{X}{1}}^T ( \bm{L} \otimes \bm{I}_{n_1 n_2} ) \tvec{\tfold{X}{1}} \nonumber \\
&=&  \fnorm{ ( \bm{L}^{1/2} \otimes \bm{I}_{n_1 n_2} ) \tvec{\tfold{X}{1}} }^2
\end{eqnarray}
since $\bm{I}^{1/2} = \bm{I}$.
Using the properties of Kronecker products above, we have
\begin{eqnarray}
( \bm{L}^{1/2} \otimes \bm{I}_{n_1 n_2} ) \tvec{\tfold{X}{1}} & = & 
( (\bm{L}^{1/2} \otimes \bm{I}_{n_2}) \otimes \bm{I}_{n_1} ) \tvec{\tfold{X}{1}} \nonumber \\ \nonumber
& = & \bm{I}_{n_1} \tfold{X}{1} ( \bm{L}^{1/2} \otimes \bm{I}_{n_2} ) \\
& = & \tensor{X} \tmmul{1} \bm{I} \tmmul{2} \bm{I} \tmmul{3} \bm{L}^{1/2}
\end{eqnarray}
Therefore,
\bgeq
\sum_{ij} c_{ij} \fnorm{\fslice{X}{i} - \fslice{X}{j}}^2 = 
\fnorm{ \tensor{X} \tmmul{1} \bm{I} \tmmul{2} \bm{I} \tmmul{3} \bm{L}^{1/2} }^2
\eneq
\QED

\subsection{Proof}

Recall that our model considers the similarity among the frontal slices of the tensor $\tensor{R}$. Using Lemma~\ref{lemma:sim3}, the objective function of our model is 
\begin{eqnarray}
\label{eq:basemodel}
f(\tensor{R}, \bm{A}) & = & \fnorm{\tensor{X}- \tensor{R} \tmmul{1} \bm{A} \tmmul{2} \bm{A}}^2 + 
\lambda_{a} \fnorm{\bm{A}}^2 + \lambda_{g} \fnorm{\tensor{R}}^2 + \nonumber \\
& & 	\lambda_s \fnorm{\tensor{R} \tmmul{3} \bm{S} }^2 
\end{eqnarray}
where $\bm{S} = (\mrm{deg}(\bm{C})-\bm{C})^{1/2}$ and  $\bms{\lambda} \geq \bm{0}$.
Since $f$ is of degree 4 in $\bm{A}$, it will difficult to optimize it efficiently. We employ the split--variable trick~\cite{nickel2013tensor,parikh2014proximal}, by splitting the variable matrix 
$\bm{A}$ into a tensor $\tensor{A}$  with exactly two frontal square slices $\bm{A}_1$ and $\bm{A}_2$
and enforce the constraint $\fnorm{\bm{A}_{1}-\bm{A}_{2}}^2=0$,  to obtain the objective function
\begin{eqnarray}
\hspace{-0.5cm}f'(\tensor{R}, \bm{A}_{1}, \bm{A}_{2})& = &\fnorm{\tensor{X}- \tensor{R} \tmmul{1} \bm{A}_{1} \tmmul{2} \bm{A}_{2}}^2  + \lambda_{a} \fnorm{\bm{A}_{1}}^2  + \lambda_{a} \fnorm{\bm{A}_{2}}^2 \nonumber\\ 
&  &  + \lambda_{g} \fnorm{\tensor{R}}^2 + \lambda_s \fnorm{\tensor{R} \tmmul{3} \bm{S} }^2   + \lambda_e \fnorm{\bm{A}_{1}-\bm{A}_{2}}^2 
\end{eqnarray}
Upon finding a minimizer $(\tensor{R}, \bm{A}_{1}, \bm{A}_{2})$ of $f'$, we use the point 
$(\tensor{R}, (\bm{A}_1 + \bm{A}_2)/2)$ for our original objective function $f$.

Because the Frobenius norm is convex, convexity is preserved under 
affine transformation, and the sum of convex functions is convex, it follows that our function 
$f'(\tensor{R},  \bm{A}_{1}, \bm{A}_{2})$ is only separately convex with respect to $\tensor{R}$, $\bm{A}_1$, and $\bm{A}_2$ 
(i.e block separately convex).
Unfortunately, $f'$ is not strictly convex with respect to these three block arguments, and is neither separately 
convex in just two blocks,   
which may lead to ALS not converging 
when trying to optimize $f'$.
Notice that ALS is essentially block-based Gauss--Seidel with 3 blocks of unknowns/variables, the blocks
$\tensor{R}$, $\bm{A}_1$, and $\bm{A}_2$, since optimizing for one block while 
keeping the other two blocks fixed is a least--squares problem.

We seek to avoid non-convergence by constructing a modified objective function 
$\hat{f}_{\rho}(\tensor{R},  \bm{A}_{1}, \bm{A}_{2})$ 
such that \\
$\lim_{\rho \rightarrow \infty} \hat{f}_{\rho}(\tensor{R}, \bm{A}_{1}, \bm{A}_{2}) = f'(\tensor{R},  \bm{A}_{1}, \bm{A}_{2})$, and 
for which block--based Gauss--Seidel is guaranteed to converge to minimizer of $\hat{f}_{\rho}$ for each $\rho$.
To this end, we modify $f'$ to make it strictly convex with respect to each of three blocks of unknowns
$\tensor{R}, \bm{A}_1, \bm{A}_2$.  
In particular, we employ a trick used in proximal algorithms \cite{parikh2014proximal} (utilizing the fact that 
strict convexity  is retained upon addition with a convex function), and 
add for each block a strictly convex term for that block that goes to 0 as $\rho \rightarrow \infty$.
\begin{eqnarray}
\hat{f}_{\rho}(\tensor{R},  \bm{A}_{1}, \bm{A}_{2}) = f'(\tensor{R},  \bm{A}_{1}, \bm{A}_{2}) + 
\frac{1}{\rho} ( \fnorm{\tensor{R}}^2 + \fnorm{\bm{A}_{1}}^2+ \fnorm{\bm{A}_{2}}^2)
\end{eqnarray}
Note that the block--partial Hessians of the added term is $\bm{I}/\rho$ 
which is positive definite and hence strictly convex. The block-gauss siedle (which coincides with ALS in this case) for $prox_f$ converges to a critical point \cite{grippof1999globally}(Th. 6.1.6.2,6.3)
At the same time, for each $\hat{f}$, $\rho \rightarrow \infty$, a critical point in $\hat{f}$ converges to a critical point in $f$  as $\hat{f}$ is continuous in $\rho$. Hence, we have an algorithm for finding a critical point of $f$. The complexity of finding a critical point of $\hat{f}$ is same as the ALASLAN for REGAL.

\end{document}